%% file: 0-main.tex
\documentclass[11pt,a4paper]{article}
\usepackage[hyperref]{emnlp2020}
\usepackage{times}
\usepackage{latexsym}

\usepackage{microtype}
\usepackage{hyperref}       % hyperlinks
\usepackage{url}            % simple URL typesetting
\usepackage{booktabs}       % professional-quality tables
\usepackage{amsfonts}       % blackboard math symbols
\usepackage{nicefrac}       % compact symbols for 1/2, etc.
\usepackage{microtype}      % microtypography
\usepackage[linesnumbered,algoruled,boxed,noend]{algorithm2e}
\usepackage{multirow}
\usepackage{subfig}
\usepackage{xspace}
\usepackage{graphicx,wrapfig}
\usepackage{amsmath}
\usepackage{xcolor}
\usepackage{enumitem}

\newcommand{\bH}{\boldsymbol{H}}
\newcommand{\bv}{\mathbf{v}}

\newcommand{\bh}{\boldsymbol{h}}
\newcommand{\bbh}{\mathbf{h}}

\newcommand{\bW}{\boldsymbol{W}}
\newcommand{\bw}{\boldsymbol{w}}
\newcommand{\mbR}{\mathbb{R}}
\newcommand{\prob}{p}

\newcommand{\calE}{G}

\SetKwInput{KwInput}{Input}
\SetKwInput{KwOutput}{Output}
\SetEndCharOfAlgoLine{}

\newcommand{\method}{\textsc{RE-Net}\xspace}
\newcommand{\para}[1]{\smallskip\noindent\textbf{#1}}

\newcommand{\teN}{\text{N}}
\newcommand{\bs}{\mathrm{s}}
\newcommand{\br}{\mathrm{r}}
\newcommand{\bo}{\mathrm{o}}
\newcommand{\bt}{t}
\newcommand{\es}{\boldsymbol{e}_{\mathrm{s}}}
\newcommand{\er}{\boldsymbol{e}_{\mathrm{r}}}
\newcommand{\eo}{\boldsymbol{e}_{\mathrm{o}}}
\renewcommand{\bh}{\boldsymbol{h}}
\newcommand{\nop}[1]{}
\newcommand{\eg}{{e.g.}}
\newcommand{\ie}{{i.e.}}

\usepackage{nccmath}

\aclfinalcopy % Uncomment this line for the final submission
\title{Recurrent Event Network: Autoregressive Structure Inference over Temporal Knowledge Graphs}

\author{Woojeong Jin$^1$ \quad Meng Qu$^2\, ^3$ \quad Xisen Jin$^1$ \quad Xiang Ren$^1$ \\
$^1$Department of Computer Science, University of Southern California\\
$^2$MILA - Quebec AI Institute \\
$^3$University of Montr\'{e}al\\
\texttt{\{woojeong.jin,xisenjin,xiangren\}@usc.edu meng.qu@umontreal.ca} \\
}

\date{}

\begin{document}
\maketitle
\begin{abstract}
\input{000abs}

\end{abstract}

\section{Introduction}

\input{010intro}

\input{020method}

\section{Experiments}

\input{030experiments}

\section{Related Work}
\input{040related}

\section{Conclusion}
\input{060conclusion}

\section*{Acknowledgement}
This research is based upon work supported in part by the Office of the Director of National Intelligence (ODNI), Intelligence Advanced Research Projects Activity (IARPA), via Contract No. 2019-19051600007, the DARPA MCS program under Contract No. N660011924033 with the United States Office Of Naval Research, the Defense Advanced Research Projects Agency with award W911NF-19-20271, and NSF SMA 18-29268. The views and conclusions contained herein are those of the authors and should not be interpreted as necessarily representing the official policies, either expressed or implied, of ODNI, IARPA, or the U.S. Government. We would like to thank all the collaborators in USC INK research lab for their constructive feedback on the work.

\bibliographystyle{acl_natbib}
\bibliography{biblio}

\clearpage
\appendix

\input{appendix}

\end{document}

%% file: 000abs.tex
% Modeling dynamically-evolving, multi-relational graph data has received a surge of interest with the rapid growth of heterogeneous event data. 
% Predicting future events (or interactions) in an autoregressive fashion is challenging, i.e., predicting events in a future time steps requires previously predicted (or observed) events.
% Predicting future interactions on multi-relational graph data requires structure inference over time and the ability to integrate temporal and structural information.

% Reasoning on temporal knowledge graphs 
% requires temporal and structural information structure inference over time and integration of 

% Reasoning on knowledge graphs over time requires forecasting ability on unseen time stamps.

Knowledge graph reasoning is a critical task in natural language processing.
The task becomes more challenging on temporal knowledge graphs, where each fact is associated with a timestamp.
Most existing methods focus on reasoning at past timestamps and they are not able to predict facts happening in the future.
This paper proposes Recurrent Event Network (\method), a novel autoregressive architecture for predicting future interactions.
The occurrence of a fact (event) is modeled as a probability distribution conditioned on temporal sequences of past knowledge graphs. 
Specifically, our \method employs a recurrent event encoder to encode past facts, and uses a neighborhood aggregator to model the connection of facts at the same timestamp.
Future facts can then be inferred in a sequential manner based on the two modules.
We evaluate our proposed method via link prediction at future times on five public datasets. Through extensive experiments, we demonstrate the strength of \method, especially on multi-step inference over future timestamps, and achieve state-of-the-art performance on all five datasets\footnote{\small \url{https://github.com/INK-USC/RE-Net}}.

% \textcolor{blue}{Meng: Knowledge graph reasoning is a critical task in natural language processing. The task becomes more challenging on temporal knowledge graphs, where each fact is associated with a timestamp. Most existing methods focus on reasoning at past timestamps, which are not able to predict facts happening in the future. This paper proposes Recurrent Event Network (\method), a novel autoregressive architecture for predicting future interactions. 
% A \method employs a recurrent event encoder to encode past facts, and uses a neighborhood aggregator to model the connection of facts at the same timestamp.
% Future facts can then be inferred in a sequential manner based on the two modules. We evaluate our proposed method via link prediction at future times on five public datasets. Extensive experiments\footnote{\small Code and data has been uploaded.} demonstrate the strength of \method, especially on multi-step inference over future time stamps. }

%

%% file: 010intro.tex
\begin{figure}[tb!]
    \centering
        {\includegraphics[width=1.0\columnwidth]{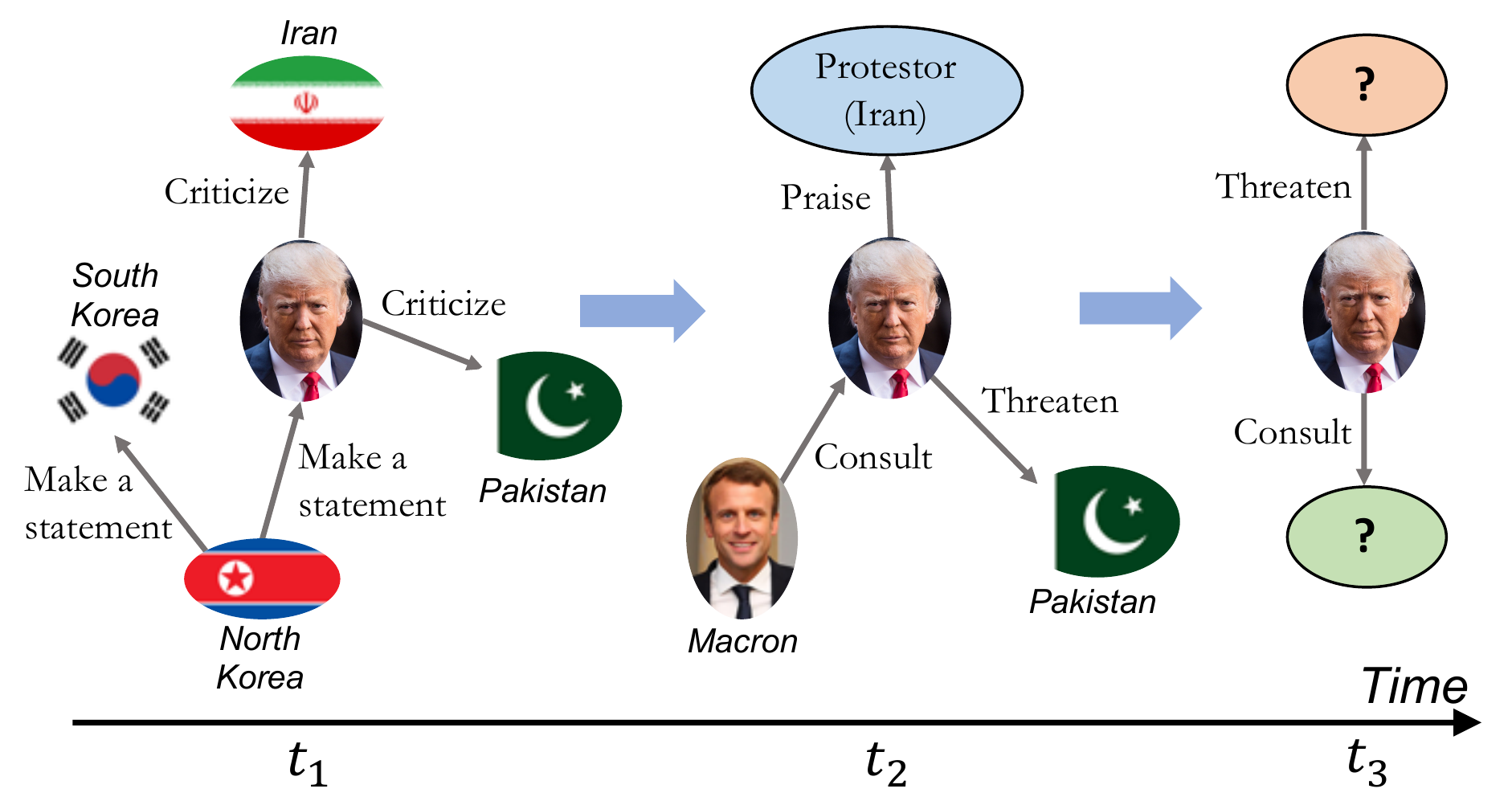}}
        \caption{\textbf{Example temporal knowledge subgraphs.} 
        Each edge or interaction between entities is associated with temporal information and a set of interactions build a multi-relational graph at each time. Our task is to predict interactions and build graphs at future times. 
        }
        \label{fig:data}
% \vspace{-0.2cm}
\end{figure}

Knowledge graphs (KGs), which store real-world facts, are vital in various natural language processing applications~\cite{Bordes2013TranslatingEF,Schlichtkrull2018ModelingRD,kazemi2019relational}. Due to the high cost of annotating facts, most knowledge graphs are far from complete, and thus predicting missing facts (a.k.a., knowledge graph reasoning) becomes an important task. Most existing efforts study reasoning on standard knowledge graphs, where each fact is represented as a triple of subject entity, object entity and the relation between them. However, in practice, each fact may not be true forever, and hence it is useful to associate each fact with a timestamp as a constraint, yielding a temporal knowledge graph (TKG). Fig.~\ref{fig:data} shows example subgraphs of a temporal knowledge graph. Despite the ubiquitousness of TKGs, methods for reasoning over such kind of data are relatively unexplored.

Given a temporal knowledge graph with timestamps varying from $t_0$ to $t_T$, TKG reasoning primarily has two settings - \textit{interpolation} and \textit{extrapolation}. In the \textit{interpolation} setting, new facts are predicted for time $t$ such that $t_0 \leq t \leq t_T$~\cite{GarcaDurn2018LearningSE,leblay2018deriving,Dasgupta2018HyTEHT}. In contrast, \textit{extrapolation} reasoning, as a less studied setting, focuses on predicting new facts (\textit{e.g.}, unseen events) over timestamps $t$ that are greater than $t_T$ (\textit{i.e.}, $t > t_T$). The extrapolation setting is of particular interests in TKG reasoning as it helps populate the knowledge graph over future timestamps and facilitates forecasting emerging events~\cite{Muthiah2015PlannedPM,Phillips2017UsingSM,korkmaz2015combining}. 
% Hence, in this work, we address the  \textit{extrapolation} setting.

%In this paper, we address the \textit{extrapolation} problem for TKG reasoning, which involves predicting facts over multiple future timestamps.
% time $t$ such that $t_0 \leq t \leq t_T$, where inference over multiple timestamps is crucial.
% We aim to infer graph structures over multiple timestamps

Recent attempts to solve the extrapolation TKG reasoning problem are Know-Evolve~\citep{Trivedi2017KnowEvolveDT} and its extension DyRep~\citep{Trivedi2019DyRepLR}, which predict future events assuming ground truths of the preceding events are \textit{given} at inference time.
As a result, these methods are unable to predict events sequentially \textit{over future timestamps} without ground truths of the preceding events--\textit{i.e.}, a practical requirement when deploying such reasoning systems for event forecasting~\cite{ijcai2019SAGE}.
Moreover, these approaches do not model \textit{concurrent events} occurring within the same time window (\eg, a day, or 12 hours), despite their prevalence in real-world event data~\cite{boschee2015icews,leetaru2013gdelt}.
%, especially when the event timestamps are ``discrete" (versus ``continuous")---which often happens when the time window is not small enough (\textit{e.g.}, a day).
% especially when the time window is not small enough (e.g., a day).
% especially when event timestamps are discretely represented.
Thus, it is desirable to have a principled method that can extrapolate graph structures over future timestamps by modeling the concurrent events within a time window as a \textit{local} graph.

To this end, we propose an autoregressive architecture, called Recurrent Event Network (\method), for modeling temporal knowledge graphs.
% to address the extrapolation problem. 
% The key idea of our approach is to learn temporal dependency from the sequence of graphs and local structural dependency from neighborhood.
% Intuitively, we approach the problem as a sequence prediction task; our method infers future graphs in a sequential way given a sequence of preceding knowledge graphs.
% The intuition of our method is that we frame temporal knowledge graphs into a sequence of knowledge graphs.
Key ideas of \method are based on:
(1) predicting future events over multiple time stamps can be formulated as a sequential and multi-step inference problem;
% of multi-relational interactions between two entities;
(2) temporally adjacent events may carry related semantics and informative patterns, which can further help predict future events (\ie, \textit{temporal information}); 
and (3) multiple events may co-occur within the same time window and exhibit structural dependencies between entities (\ie, \textit{local graph structural information}).

\begin{figure}[tb!]
    \centering
        {\includegraphics[width=1\columnwidth]{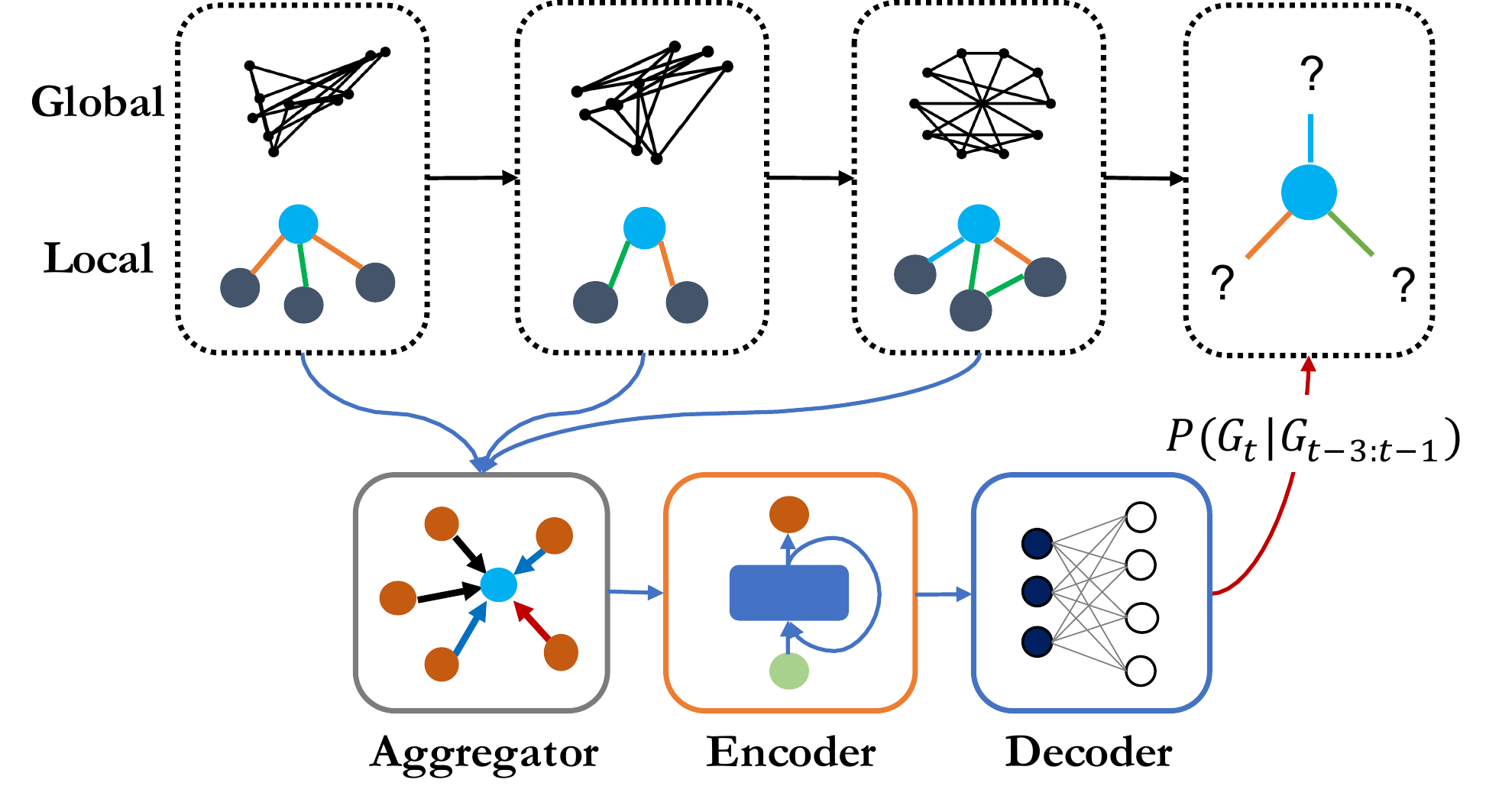} }
        \vspace{-5mm}
        \caption{\textbf{Illustration of the Recurrent Event Network architecture.}
        The aggregator encodes the global graph structure and the local neighborhood, capturing global and local information respectively. The recurrent event encoder updates its state with the sequence of encoded representations of graph structures. 
        % from the aggregator in an autoregressive manner.
        The MLP decoder defines the probability of a current graph. 
        % $\prob(\calE_t|\calE_{:t-1})$.
        % , which is conditioned on the preceding events.
        % \method employs an RNN to capture $\bs$-related interactions $\teN_t^{(\bs)}$ (modeled by a neighborhood aggregator) at different time $t$. Global information from $\calE_t$ is used to capture the global graph structures. Recurrent event encoder updates its state with graph sequences in an autoregressive manner. 
        % The decoder defines the probability $\prob(\bs_t, \br_t, \bo_t|\calE_{:t-1})$ at current time step, which is conditioned on the preceding events. 
        % \wj{Expand to have more details}
        }
        \label{fig:renet}
% \vspace{-0.3cm}
\end{figure}

Given these observations, \method defines the joint probability distribution of all events in a TKG in an autoregressive fashion.
% as a sequence of graphs, where
The probability distribution of the concurrent events at the current time step is conditioned on all the preceding events (see Fig.~\ref{fig:renet} for an illustration). 
Specifically, a \textit{recurrent event encoder} summarizes information of the past event sequences, and a \textit{neighborhood aggregator} aggregates the information of concurrent events within the same time window.
With the summarized information, our decoder defines the joint probability of a current event.
% Such an autoregressive model can be effectively trained by using teacher forcing, and
Inference for predicting future events can be achieved by sampling graphs over time in a sequential manner.

We evaluate our proposed method on five public TKG datasets via a temporal (extrapolation) link prediction task, by testing the performance of multi-step inference over time. 
Experimental results demonstrate that \method outperforms state-of-the-art models of both static and temporal knowledge graph reasoning, showing its better capability to model temporal, multi-relational graph data.
We further show that \method can perform effective multi-step inference to predict unseen entity relationships in a distant future. 

%% file: 020method.tex
\section{Problem Formulation}
\label{sec:tkg}

We first describe notations for building our model and problem definition, and then we define the joint distribution of temporal events. 
% In the next section, we will describe details of \method, our autoregressive structure inference framework. 

\para{Notations and Problem Definition.}
We consider a \textit{temporal knowledge graph} as a multi-relational, directed graph with time-stamped edges between nodes (entities).
An \textit{event} is defined as a time-stamped edge, \textit{i.e.}, (subject entity, relation, object entity, time) and is denoted by a quadruple $(\bs,\br,\bo,t)$ or $(\bs_t, \br_t, \bo_t)$. We denote a set of events at time $t$ as ${\calE}_t$. 
In our setup, the timestamps are discrete integers and used for the relative order of graphs or events.
A TKG is built upon \textit{a sequence of event quadruples} ordered ascending based on their timestamps, \textit{i.e.}, $\{\calE_{\bt}\}_t = \{(\bs_i,\br_i,\bo_i,t_i)\}_i$ ($t_i < t_j, \forall i<j$), 
where each time-stamped edge has a direction pointing from the subject entity to the object entity.\footnote{The same triple $(\bs, \br, \bo)$ may occur multiple times in different timestamps, yielding different event quadruples.}  
The goal of \textit{learning generative models of events} is to learn a distribution $\prob(\calE)$ over TKGs, based on a set of observed event sets $\{\calE_1, ..., \calE_t\}$.

\para{Approach Overview.}
The key idea of our approach is to learn temporal dependency from the sequence of graphs and local structural dependency from the neighborhood (Fig.~\ref{fig:renet}). 
Formally, we represent TKGs as sequences, and then build an autoregressive generative model on the sequences.
To this end, \method defines the joint probability of concurrent events (or a graph), which is conditioned on all the previous events.
Specifically, \method consists of a Recurrent Neural Network (RNN) as a recurrent event encoding module and a neighborhood aggregation module to capture the information of graph structures.
We first start with the definition of joint distribution of temporal events.

\para{Modeling Joint Distribution of Temporal Events.}
We define the joint distribution of all the events $\calE = \{\calE_{1}, ..., \calE_{T}\}$ in an autoregressive manner.
% , i.e., $\prob(\calE) = \prod_{t=1}^T{\prob(\calE_t|\calE_{t-m:t-1})}$. 
Basically, we decompose the joint distribution into a sequence of conditional distributions, $\prob(\calE_t|\calE_{t-m:t-1})$), where 
we assume the probability of the events at a time step, $\calE_t$, depends on the events at the previous $m$ steps, $\calE_{t-m:t-1}$. 
For each conditional distribution $\prob(\calE_t|\calE_{t-m:t-1})$, we further assume that the events in $\calE_t$ are mutually independent given the previous events $\calE_{t-m:t-1}$. In this way, the joint distribution can be rewritten as follows:

\small
\begin{multline}
\label{eq:prob_event_t}
    \prob(\calE)= \prod_{t}\prod_{(\bs_t,\br_t,\bo_t)\in\calE_{t}}  \prob(\bs_t,\br_t,\bo_t|\calE_{t-m:t-1})\\
    = \prod_t \prod_{(\bs_t,\br_t,\bo_t)\in\calE_t} 
    \prob(\bs_t |\calE_{t-m:t-1})
    \cdot \prob(\br_t |\bs_t, \calE_{t-m:t-1})\\
    \cdot \prob(\bo_t|\bs_t,\br_t,\calE_{t-m:t-1}).
\end{multline}
\normalsize

% \begin{equation}
% \label{eq:prob_event_t}
%     \small \prob(\calE)= \prod_{t}\prod_{(\bs_t,\br_t,\bo_t)\in\calE_{t}}  \prob(\bs_t,\br_t,\bo_t|\calE_{t-m:t-1}).
% \end{equation}

% We factorize the joint probability of a triplet into three conditional distribution. 
From these probabilities, we generate triplets as follows.
Given all the past events $\calE_{t-m:t-1}$, we first sample a subject entity $\bs_t$ through $\prob(\bs_t |\calE_{t-m:t-1})$. Then we generate a relation $\br_t$ with $\prob(\br_t |\bs_t, \calE_{t-m:t-1})$, and finally the object entity $\bo_t$ is generated by $\prob(\bo_t|\bs_t,\br_t,\calE_{t-m:t-1})$.\footnote{We can also first sample an object entity in this process. Details are omitted for brevity.}
% Note that the probability is factorized in the order of $\bo_t, \br_t, \bs_t$. 

% We omit it for brevity.

% In this work,
% we assume that $\prob(\bo_t|\allowbreak\bs_t,\br_t,\calE_{t-m:t-1})$ and $\prob(\br_t | \allowbreak \bs_t, \calE_{t-m:t-1})$ depend only on events that are related to $\bs$, and focus on modeling the following joint probability:
% \begin{multline}
% \label{eq:prob_factor}
%     \prob(\bs_t,\br_t,\bo_t|\calE_{t-m:t-1}) \\=  \prob(\bs_{t}|\calE_{t-m:t-1})
%     \cdot \prob(\br_{t}|\bs,\teN^{(\bs)}_{t-m:t-1})
%     \cdot \prob(\bo_{t}|\bs,\br,\teN^{(\bs)}_{t-m:t-1}),
% \end{multline}
% where $\calE_{t}$ becomes $\teN^{(\bs)}_t$ which is a set of neighboring entities interacted with subject entity $\bs$ under \textit{all} relations at time $t$. 
% For $\prob(\bs_{t}|\calE_{t-m:t-1})$, all event sets should be considered since subject $\bs$ is not given. 
Next, we introduce how these probabilities are defined and parameterized in our method.

\section{Recurrent Event Network}
\label{sec:re-net}
In this section, we introduce our proposed method, Recurrent Event Network (\method).
\method consists of a Recurrent Neural Network (RNN) as a recurrent event encoder (Sec.~\ref{sec:encoder}) for temporal dependency and a neighborhood aggregator (Sec.~\ref{sec:aggre}) for graph structural dependency.
We also discuss parameter learning of \method and define multi-step inference for distant future by sampling intermediate graphs in a sequential manner (Sec.~\ref{sec:learn}). 

% \xiang{move and fill up some sentences to transit from previous section to here, and overview the subsections in this part.}

\subsection{Recurrent Event Encoder}
\label{sec:encoder}

To parameterize the probability for each event, \method introduces a set of \textit{global} representations as well as \textit{local} representations. The global representation $\bH_{t}$ summarizes the global information from the entire graph until time stamp $t$, which reflects the global preference on the upcoming events. In contrast, the local representations focus more on each subject entity $\bs$ or each pair of subject entity and relation $(\bs,\br)$, which capture the knowledge specifically related to those entities and relations. 
We denote the above local representations as $\bh_{t}(\bs)$ and $\bh_{t}(\bs,\br)$, respectively. 
The global and local representations capture different aspects of knowledge from the knowledge graph, which are naturally complementary, allowing us to model the generative process of the graph in a more effective way.

Based on the above representations, \method parameterizes $\prob(\bo_t| \allowbreak \bs,\br,\calE_{t-m:t-1})$ in the following way:
\begin{equation}
\label{eq:prob}
	%\prob_{\bo|\mathbf{s,r,\calE}}( \bo_{t}|\bs,\br,\teN^{(\bs)}_{t-m:t-1})=f_1(\es,\er,\bh_{t-1}(\bs, \br))[\bo_{t}],
	\small \prob( \bo_{t}|\bs,\br,\calE_{t-m:t-1}) \propto \exp \left([\es:\er:\bh_{t-1}(\bs, \br)]^\top \cdot  \bw_{\bo_{t}}\right),
\end{equation}
where $\es, \er \in \mbR^d$ are learnable embedding vectors specified for subject entity $\bs$ and relation $\br$. 
$\bh_{t-1}(\bs,\br)\in \mbR^d$ is the local representation for $(\bs,\br)$ obtained at time stamp $(t-1)$. Intuitively, $\es$ and $\er$ can be understood as static embedding vectors for subject entity $\bs$ and relation $\br$, whereas $\bh_{t-1}(\bs,\br)$ is dynamically updated at each time stamp. By concatenating both the static and dynamic representations, \method can effectively capture the semantic of $(\bs,\br)$ up to time stamp $(t-1)$. Based on that, we further compute the probability of different object entities $\bo_{t}$ by passing the encoding into our multi-layer perceptron (MLP) decoder.
We define the MLP decoder as a linear softmax classifier parameterized by $\{\bw_{\bo_{t}}\}$.

Similarly, we define probabilities for relations and subjects as follows:
% \meng{It is the first time that $H$ appear in the paper. We might need to add more explanation. Also, for all the vectors, we may use mathbf to represent them.}

\small
\begin{align}
\label{eq:prob2}
	\prob(\br_{t}|\bs,\calE_{t-m:t-1})&\propto \exp \left([\es:\bh_{t-1}(\bs))]^\top \cdot \bw_{\br_{t}} \right),\\
	\label{eq:prob3}
	\prob(\bs_{t}|\calE_{t-m:t-1})& \propto \exp \left( \bH_{t-1}^\top \cdot \bw_{\bs_{t}} \right),
\end{align}
\normalsize
where $\bh_{t-1}(\bs)$ focuses on the local information about $\bs$ in the past, and $\bH_{t-1} \in \mbR^d$ is a vector representation to encode global graph structures $\calE_{t-1:t-m}$. To predict what relations a subject entity will interact with  $\prob(\br_{t}|\bs,\calE_{t-m:t-1})$, we treat the static representation $\es$ as well as the dynamic representation $\bh_{t-1}(\bs)$ as features, and feed them into a multi-layer perceptron (MLP) decoder parameterized by $\bw_{\br_{t}}$. Besides, to predict the distribution of subject entities at time stamp $t$ (i.e., $\prob(\bs_{t}|\calE_{t-m:t-1})$), we treat the global representation $\bH_{t-1}$ as a feature, as it summarizes the global information from all the past graphs until time stamp $t-1$, which reflects the global preference on the upcoming events at time stamp $t$.

% Both the local representations $\bh_{t-1}(\bs, \br), \bh_{t-1}(\bs)$ and the global representation $\bH_{t-1}$ are dynamic, meaning that we will update them according to the observed events at each time stamp. 

The global representation $\bH_t$ is expected to preserve the global information about all the graphs up to time stamp $t$.
The local representations $\bh_t(\bs, \br)$ and $\bh_t(\bs)$ emphasize more on the local events related to each entity and relation.
Thus we define them as follows:
% We define the global representation $\bH_t$ and local representations $\bh_{t}(\bs,\br)$ and $\bh_{t}(\bs)$ as follows:
% \begin{equation}
% 	\label{eq:globalh}
% 	\small \bH_{t} = \textrm{RNN}^1(g(\calE_t), \bH_{t-1}).
% \end{equation}

\small
\begin{align}
    \bH_{t} &= \textrm{RNN}^1(g(\calE_t), \bH_{t-1}), \label{eq:globalh}\\
	\bh_{t}(\bs,\br) &= \textrm{RNN}^2(g(\teN^{(\bs)}_{t}), \bH_{t}, \bh_{t-1}(\bs,\br)),\\
	\bh_{t}(\bs) &= \textrm{RNN}^3(g(\teN^{(\bs)}_{t}), \bH_{t},\bh_{t-1}(\bs)),
\end{align}
\normalsize
where $g$ is an aggregate function which will be discussed in Section~\ref{sec:aggre} and $\teN^{(\bs)}_{t}$ stands for all the events related to $\bs$ at the current time step $t$.
We leverage a recurrent model $\textrm{RNN}$~\cite{Cho2014LearningPR} to update them.
% \footnote{Details are described in Section~\ref{append:eventseq} of appendix.}
The global representation takes the global graph structure $g(\calE_t)$ as an input.
$g(\calE_t)$ is an aggregation over all the events $\calE_t$ at time $t$. 
We define {\small$g(\calE_t) = \max(\{g(\teN_t^{(s)})\}_s)$}, which is an element-wise max-pooling operation over all {\small$g(\teN^{(\bs)}_{t})$}. The $g(\teN^{(\bs)}_{t})$ captures the local graph structure for subject entity $s$. 
The local representations are different from the global representations in two ways.
First, the local representations focus more on each entity and relation, and hence we aggregate information from events $\teN^{(\bs)}_{t}$ that are related to the entity. Second, to allow \method to better characterize the relationships between different entities, we treat the global representation $\bH_{t}$ as an extra feature in the definition, which acts as a bridge to connect different entities.

In the next section, we introduce how we design $g$ in \method.
% The update rules require a proper aggregation function $g$ to effectively encode events. However, it is nontrivial because for each subject entity $\bs$, it can interact with multiple relations and object entities at each time step $t$, \ie, the set {\small$\teN^{(\bs)}_{t}$} can contain multiple events. Next, we introduce how we design $g$ in \method.

\begin{figure}[tb!]
    \centering
    \includegraphics[width=1\columnwidth]{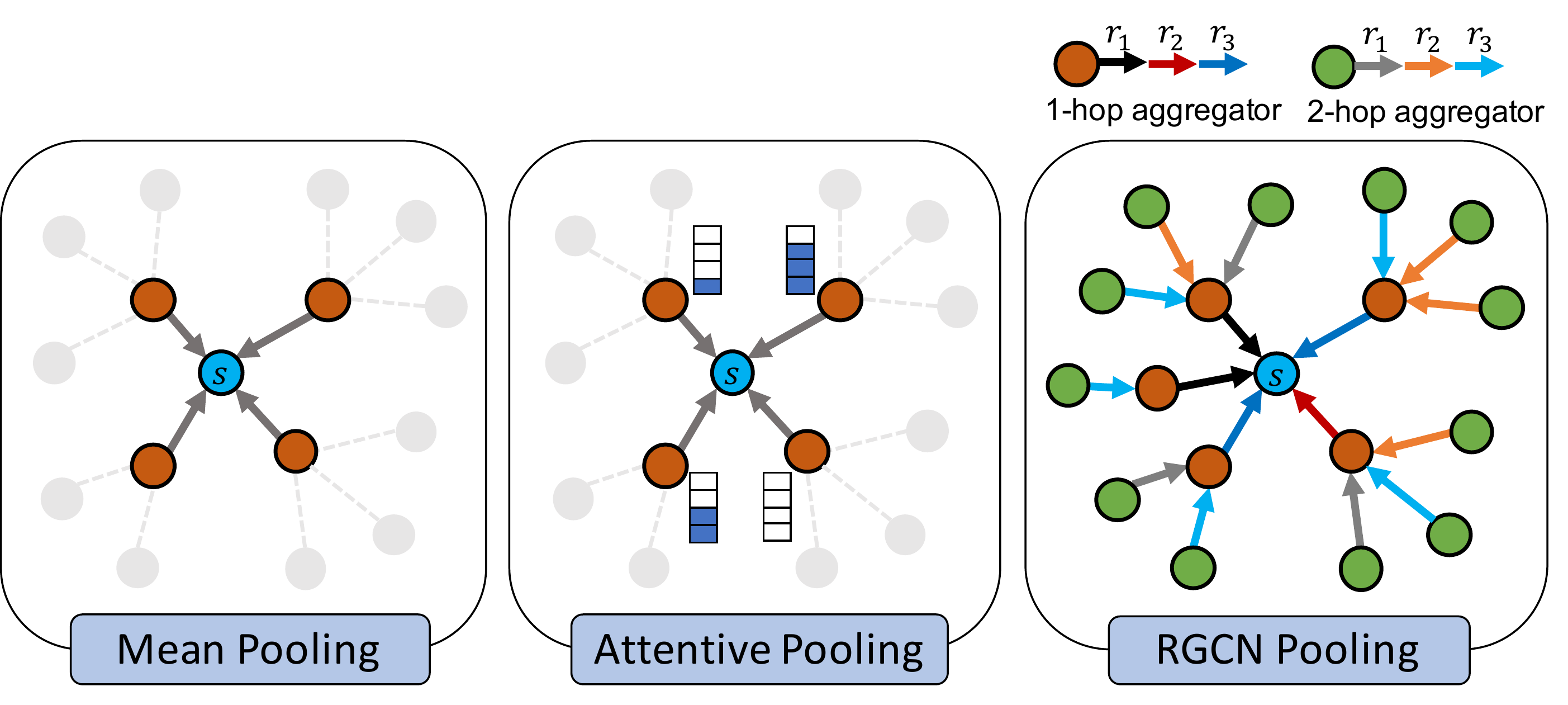}
    \caption{\textbf{Comparison of neighborhood aggregators.}
    The blue node corresponds to node $\bs$, red nodes are 1-hop neighbors, and green nodes are 2-hop neighbors. Different colored edges are different relations.
    Mean and attentive pooling aggregators do not differentiate different relations and do not encode 2-hop neighbors, whereas RGCN aggregator can incorporate information from multi-relational and multi-hop neighbors.
    }
    \label{fig:rgcnaggre}
\end{figure}

\subsection{Neighborhood Aggregators}
\label{sec:aggre}
% \meng{It might be better if we can explain the high-level intuition here. Otherwise, readers may not fully understand the motivation of the RGCN aggregator.}
% Recall that \method relies on an aggregation function $g$, which aims at summarizing the local information and the global information. 
In this section, we first introduce two simple aggregation functions: a mean pooling aggregator and an attentive pooling aggregator. These two simple aggregators only collect neighboring entities under the same relation $\br$. Then we introduce a more powerful aggregation function: a multi-relational aggregator.
We depict comparison on aggregators in Fig.~\ref{fig:rgcnaggre}.
% \textcolor{red}{Yu: IMO no need to italicize i.e. \& e.g. as they are common latin abbrv in english}
%Before going into a multi-relational graph aggregator, we first introduce a simple mean. 
% We assume that the next set of objects can be predicted with the previous object history under the same relation. \meng{I don't quite understand the sentence.}

\para{Mean Pooling Aggregator.}
The baseline aggregator simply takes the element-wise mean of representations in {\small $\{\eo: \bo \in \teN_t^{(\bs,\br)}\}$}, where {\small$\teN_t^{(\bs,\br)}$} is the set of objects that interacted with $\bs$ under $\br$ at $t$. 
% \textcolor{red}{Yu: Minor grammar mistake - "is to" is somehow improper}
But the mean aggregator treats all neighboring objects equally, and thus ignores the different importance of each neighbor entity. 

\para{Attentive Pooling Aggregator.}
We define an attentive aggregator based on the additive attention introduced in \citep{Bahdanau2015NeuralMT} to distinguish the important entities for $(\bs,\br)$. 
The aggregate function is defined as {\small $g(\teN_t^{(\bs,\br)}) = \sum_{\bo \in \teN_t^{(\bs, \br)}} \alpha_\bo \eo$},
where {\small $\alpha_\bo = \textrm{softmax}(\bv^\top \tanh(\bW(\es; \er; \eo)))$}.
{\small  $\bv \in \mathbb{R}^d$} and  {\small $\bW \in \mathbb{R}^{d \times 3d}$} are trainable weight matrices. 
By adding the attention function of the subject and the relation, the weight can determine how relevant each object entity is to the subject and the relation.

\para{Multi-Relational Graph (RGCN) Aggregator.} 
% \textcolor{red}{Yu: Is this the identical aggregator appeared in R-GCN paper?}
We introduce a multi-relational graph aggregator from \citep{Schlichtkrull2018ModelingRD}. 
This is a general aggregator that can incorporate information from multi-relational and multi-hop neighbors. Formally, the aggregator is defined as follows:
\begin{equation}
\label{eq:rgcn}
    \small g(\teN^{(\bs)}_t)= \bbh_s^{(l+1)} = \sigma\Big(\sum_{\br\in R}\sum_{\bo \in \teN_t^{(\bs,\br)}} \frac{1}{c_s} \bW_r^{(l)} \bbh_\bo^{(l)} + \bW_0^{(l)} \bbh_s^{(l)}\Big), 
\end{equation}
where initial hidden representations for each node ($\bbh_o^{(0)}$) are set to trainable embedding vectors ($\eo$) for each node and $c_s$ is a normalizing factor.
Details are described in Section~\ref{sec:rgcn} of appendix.

% \wj{I moved details to appendix.}

\begin{algorithm}[t]
\begin{small}
    \caption{\small Learning algorithm of \method}\label{alg:renet}
	\KwInput{Observed graph sequence: $\{\calE_{1}, ..., \calE_{t}\}$, Number of events to sample at each step: $M$.}
	\KwOutput{An estimation of the conditional distribution $\prob(\calE_{t+\Delta t}|\calE_{:t})$.}
	$t' \leftarrow t+1$\\
	\While{$t' \leq t+ \Delta t$}{
    Sample $M$ number of $\bs ~\sim \prob(\bs|\hat{\calE}_{t+1:t'-1},\calE_{:t})$ by equation \eqref{eq:prob3}.\label{alg:samples}\\
    Pick top-$k$ triples $\{(\bs_1,\br_1,\bo_1, t'),...,(\bs_k,\br_k,\bo_k, t')\}$ ranked by $\prob(\bs,\br,\bo|\hat{\calE}_{t+1:t'-1},\calE_{:t})$. \label{alg:sampletriple}\\
    ${\hat{G}}_{t'} \leftarrow \{(\bs_1,\br_1,\bo_1, t'),...,(\bs_k,\br_k,\bo_k, t')\}$ \label{alg:graph}\\
	$t' \leftarrow t'+ 1$
	}
	Estimate the probability of each event $\prob(\bs,\br,\bo|\hat{\calE}_{t+1:t+\Delta t-1},\calE_{:t})$.\\
	Estimate the joint distribution of all events $\prob(\calE_{t+\Delta t}|\hat{\calE}_{t+1:t+\Delta t-1},\calE_{:t})$ by equation \eqref{eq:prob_event_t}.
	\\
	\Return $\prob(\calE_{t+\Delta t}|\hat{\calE}_{t+1:t+\Delta t-1},\calE_{:t})$ as an estimation.
	\end{small}
\end{algorithm}

\subsection{Parameter Learning and Inference}
\label{sec:learn}
In this section, we discuss how \method is trained and infers events over multiple time stamps.

\para{Parameter Learning via Event Predictions.}
An object entity prediction given $(\bs,\br)$ can be viewed as a multi-class classification task, where each class corresponds to each object entity.
Similarly, relation prediction given $\bs$ and subject entity prediction can be considered as a multi-class classification task. 
Here we omit the notations for preceding events for brevity.
% To learn parameters and representations of entities and relations, we adopt a multi-class cross-entropy loss to the model's output.
Thus, the loss function is as follows:
% The loss function is comprised of three losses and is defined as:
\begin{equation}
    \small \mathcal{L} = -\sum_{(\bs,\br,\bo,t) \in \calE}  \log\prob(\bo_t|\bs_t,\br_t) +\lambda_1 \log\prob(\br_t|\bs_t) + \lambda_2 \log\prob(\bs_t),
    \label{loss}
\end{equation}
where $\calE$ is set of events, and $\lambda_1$ and $\lambda_2$ are importance parameters that control the importance of each loss term. $\lambda_1$ and $\lambda_2$ can be chosen depending on the task. 
If a task aims to predict  $o$ given $(s,r)$, then we can give small values to $\lambda_1$ and $\lambda_2$.
% Each probability in equation~\eqref{loss} is defined in equations \eqref{eq:prob}, \eqref{eq:prob2}, and \eqref{eq:prob3}, respectively.
% We apply teacher forcing for model training over historical data, i.e., we use the ground truth rather than the model's own prediction as the input of the next time step during training.

\para{Multi-step Inference over Time.} 
\method seeks to predict the forthcoming events based on the previous observations. 
Suppose that the current time is $t$ and we aim to predict events at time $t + \Delta t$ where $\Delta t>0$. 
Then the problem of multi-step inference can be formalized as inferring the conditional probability {\small $\prob (\calE_{t+\Delta t}|\calE_{:t})$}. 
The problem is nontrivial as we need to integrate over all $\calE_{t+1:t+\Delta t-1}$. To achieve efficient inference, we draw a sample of $\calE_{t+1:t+\Delta t-1}$, and estimate the conditional probability as follows:

\small
\begin{align*}
	&\prob (\calE_{t+\Delta t}|\calE_{:t})\nonumber\\
	&= \sum_{\calE_{t+1:t+\Delta t-1}} \prob (\calE_{t+\Delta t}, \calE_{t+1:t+\Delta t-1}|\calE_{:t})\\ 
	&= \sum_{\calE_{t+1:t+\Delta t-1}} \prob(\calE_{t+\Delta t}|\calE_{:t+\Delta t-1})  \cdots  \prob(\calE_{t+1}|\calE_{:t})\\ 
	&= \mathbb{E}_{\calE_{t+1:t+\Delta t-1}|\calE_{:t}} [\prob(\calE_{t+\Delta t}|\calE_{:t+\Delta t-1})] \\
	&\simeq \prob(\calE_{t+\Delta t}|\hat{\calE}_{t+1:t+\Delta t-1},\calE_{:t}).
\end{align*}
\normalsize

Intuitively, one starts with computing {\small $\prob(\calE_{t+1}|\calE_{:t})$}, and drawing a sample $\hat{\calE}_{t+1}$ from the conditional distribution. With this sample, one can further compute {\small $\prob(\calE_{t+2}|\hat{\calE}_{t+1},\calE_{:t})$}. By iteratively computing the conditional distribution for $\calE_{t'}$ and drawing a sample from it, one can eventually estimate {\small $\prob (\calE_{t+\Delta t}|\calE_{:t})$} as {\small $\prob(\calE_{t+\Delta t}|\hat{\calE}_{t+1:t+\Delta t-1},\calE_{:t})$}. 
Although we can improve the estimation by drawing multiple graph samples at each step, \method already performs very well with a single sample, and thus we only draw one sample graph at each step for better efficiency. 
Based on the estimation of the conditional distribution, we can further predict events that are likely to form in the future. 
We summarize the detailed inference algorithm in Algorithm~\ref{alg:renet}; we first sample $M$ number of $\bs$ (line~\ref{alg:samples}) and pick top-$k$ triples (line~\ref{alg:sampletriple}). Then we build a graph at time $t'$ (line~\ref{alg:graph}) to generate a graph. The time complexity of the algorithm is described in Section~\ref{sec:complexity} of appendix.

%% file: 030experiments.tex
\begin{table*}[tb!]
\caption{Performance comparison on temporal link prediction (average metrics in \% over 5 runs) on three event-based TKG datasets (ICEWS18, GDELT, and ICEWS14) and two public knowledge graphs (WIKI and YAGO). \method achieves the best results.
% \footnote{Results with raw setting are in the appendix.}
}
\label{result:filtered}
\centering
\resizebox{1.0\textwidth}{!}{
\begin{tabular}{ll|ccc|ccc|ccc|ccc|ccc}
\cmidrule[1pt](){2-17}
    \multicolumn{2}{c}{\multirow{2}{*}{\vspace{-2.2mm}\hspace{-14mm}\textbf{Method}}} & \multicolumn{3}{c}{\textbf{ICEWS18}} & \multicolumn{3}{c}{\textbf{GDELT}} & \multicolumn{3}{c}{\textbf{ICEWS14}} &\multicolumn{3}{c}{\textbf{WIKI}}& \multicolumn{3}{c}{\textbf{YAGO}} \\ \cmidrule(lr){3-5} \cmidrule(lr){6-8} \cmidrule(lr){9-11} \cmidrule(lr){12-14} \cmidrule(lr){15-17}
    &                   &MRR    &H@3    &H@10   &MRR    &H@3    &H@10   &MRR    &H@3    &H@10   &MRR    &H@3    &H@10   &MRR    &H@3    &H@10  \\ 
\cmidrule{2-17}
\multirow{4}{*}{\rotatebox{90}{\hspace*{-6pt}Static}} 
&DistMult &22.16  &26.00  &42.18  &18.71  &20.05  &32.55  &19.06  &22.00  &36.41  &46.12  &49.81  &51.38  &59.47  &60.91  &65.26  \\
&R-GCN&23.19  &25.34  &36.48  &23.31  &24.94  &34.36  &26.31  &30.43  &45.34  &37.57  &39.66  &41.90  &41.30  &44.44  &52.68  \\
&ConvE&36.85  &39.92  &50.54  &35.56  &39.45  &49.16  &40.46  &43.33  &54.75  &47.55  &49.78  &49.42  &62.66  &63.36  &65.57  \\
&RotatE&23.10  &27.61  &38.72  &22.33  &23.89  &32.29  &29.56  &32.92  &42.68  &48.67  &49.74  &49.88  &64.09  &64.67  &66.16  \\
\cmidrule[0.6pt](){2-17}
\multirow{12}{*}{\rotatebox{90}{\hspace*{-6pt}Temporal}}
&TA-DistMult&28.53  &31.57  &44.96  &29.35  &31.56  &41.39  &20.78  &22.80  &35.26  &48.09  &49.51  &51.70  &61.72  &63.32  &65.19  \\ 
&HyTE&7.31   &7.50   &14.95  &6.37   &6.72   &18.63  &11.48  &13.04  &22.51  &43.02  &45.12  &49.49  &23.16  &45.74  &51.94  \\ 
% \cmidrule{2-17}
&dyngraph2vecAE&1.52   &1.99   &2.02   &4.53   &1.87   &1.87   &10.83  &12.70  &15.02  &5.30   &5.27   &5.45   &0.93   &0.84   &0.95    \\
% &DynTriad               &3.48   &3.55   &11.47  &xxx    &xxx    &xxx    &8.4    &12.45  &24.24  &2.62   &4.26   &6.63   &xxx    &xxx    &xxx    \\
&tNodeEmbed&8.32   &9.74   &17.47  &19.97  &22.62  &32.72  &17.84  &20.16  &32.88  &9.54   &10.44  &16.60  &4.22   &4.16   &8.4    \\
&EvolveRGCN&16.59  &18.32  &34.01  &15.55  &19.23  &31.54  &17.01  &18.97  &32.58  &46.49  &47.83  &49.23    &59.74  &61.03  &61.69    \\
&Know-Evolve* &3.27   &3.23   &3.26   &2.43   &2.35   &2.41   &1.42   &1.37   &1.43   &0.09   &00.03  &0.10   &00.07  &0      &0.04   \\ 
&Know-Evolve+MLP        &9.29   &9.62   &17.18  &22.78  &25.49  &35.41  &22.89  &26.68  &38.57  &12.64  &14.33  &21.57  &6.19   &6.59   &11.48  \\ 
&DyRep+MLP &9.86   &10.66  &18.66  &23.94  &27.88  &36.58  &24.61  &28.87  &39.34  &11.60  &12.74  &21.65  &5.87   &6.54   &11.98  \\ 
&R-GCRN+MLP&35.12  &38.26  &50.49  &37.29  &41.08  &51.88  &36.77  &40.15  &52.33  &47.71  &48.14  &49.66  &53.89  &56.06  &61.19  \\ 
\cmidrule{2-17}
&\method w. mean agg.   &40.70  &43.27   &53.65 &38.35  &42.13  &52.52  &43.79  &47.34  &57.47  &51.13  &51.37  &53.01  &65.10  &65.24  &67.34  \\
&\method w. attn agg.   &40.96  &44.08   &54.32 &38.54  &42.25  &52.85  &43.94  &47.85  &57.91	&51.25  &52.54  &53.12  &65.13  &65.54  &67.87  \\
&\method                &\textbf{42.93} &\textbf{45.47} &\textbf{55.80} &\textbf{40.42} &\textbf{43.40} &\textbf{53.70} &\textbf{45.71} &\textbf{49.06} &	\textbf{59.12}  &\textbf{51.97} &\textbf{52.07} &\textbf{53.91} &\textbf{65.16} &\textbf{65.63} &\textbf{68.08}\\
\cmidrule[1pt](){2-17}
\end{tabular}}
\end{table*}

Evaluating the quality of generated graphs is non-trivial, especially for knowledge graphs~\citep{theis2015note}.
In our experiments, we evaluate the proposed method on a \textit{extrapolation} link prediction task on TKGs. 
The task of predicting future links aims to predict unseen relationships with object entities given $(\bs,\br,?,t)$ (or subject entities given $(?,\br,\bo,t)$) at future time $t$, based on the past observed events in the TKG. Essentially, the task is a ranking problem over all the events $(\bs,\br,?,t)$ (or $(?,\br,\bo,t)$). \method can approach this problem by computing the probability of each event in a distant future with the inference algorithm in Algorithm~\ref{alg:renet}, and further rank all the events according to their probabilities.
Note that we are only given a training set as ground truth at inference and we do not use any ground truth in the test set for the next time step predictions when performing multi-step inference.
This is the main difference from previous work; they use previous ground truth in the test set.

We evaluate our proposed method on three benchmark tasks: (1) predicting future events on three event-based datasets; (2) predicting future facts on two knowledge graphs which include facts with time spans, and (3) studying ablation of our proposed method.
Section~\ref{exp:setup} summarizes the datasets.
% and the appendix contains additional information.
In all these experiments, we perform predictions on time stamps that are not observed during training.

\subsection{Experimental Setup}
\label{exp:setup}
We compare the performance of our model against various traditional models for knowledge graphs, as well as some recent temporal reasoning models on five public datasets.

\para{Datasets.}
We use five TKG datasets in our experiments: 1) three event-based TKGs: ICEWS18~\citep{boschee2015icews}, ICEWS14~\citep{Trivedi2017KnowEvolveDT}, and GDELT~\citep{leetaru2013gdelt}; and 2) two knowledge graphs where temporally associated facts have meta-facts as $(\bs,\br,\bo, [t_s, t_e])$ where $t_s$ is the starting time point and $t_e$ is the ending time point:  WIKI~\citep{leblay2018deriving} and YAGO~\citep{Mahdisoltani2014YAGO3AK}.
% \footnote{Details of the datasets are described in Section~\ref{append:dataset} of appendix.}

\para{Evaluation Setting and Metrics.}
For each dataset 
except ICEWS14\footnote{We used the splits as provided in \citep{Trivedi2017KnowEvolveDT}.}, 
we split it into three subsets, i.e., train(80\%)/valid(10\%)/test(10\%), by time stamps. Thus, (time stamps of train) $<$ (time stamps of valid) $<$ (time stamps of test).
% Note that this is different from the previous way which randomly picks train, valid, and test sets regardless of time stamps.
We report a filtered version of Mean Reciprocal Ranks (MRR) and Hits$@3$/$10$. 
Similar to the definition of filtered setting in \citep{Bordes2013TranslatingEF}, during evaluation, we remove all the valid triplets that appear in the train, valid, or test sets from the list of corrupted triplets.

\para{Baselines.}
We compare our approach to baselines for static graphs and temporal graphs as follows:

\para{(1) \textit{Static Methods.}}
By ignoring the edge time stamps, we construct a static, cumulative graph for all the training events, and apply multi-relational graph representation learning methods including DistMult~\citep{Yang2015EmbeddingEA}, R-GCN~\citep{Schlichtkrull2018ModelingRD}, ConvE~\citep{Dettmers2018Convolutional2K}, and RotatE~\citep{sun2019rotate}.

\begin{figure}[!tb]
    \centering
    {\includegraphics[width=0.8\columnwidth]{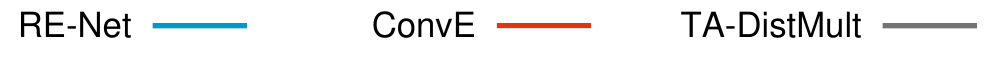}
    % \vspace{-3mm}
    }
    \\
    \subfloat[ICEWS18]{\includegraphics[width=0.46\columnwidth]{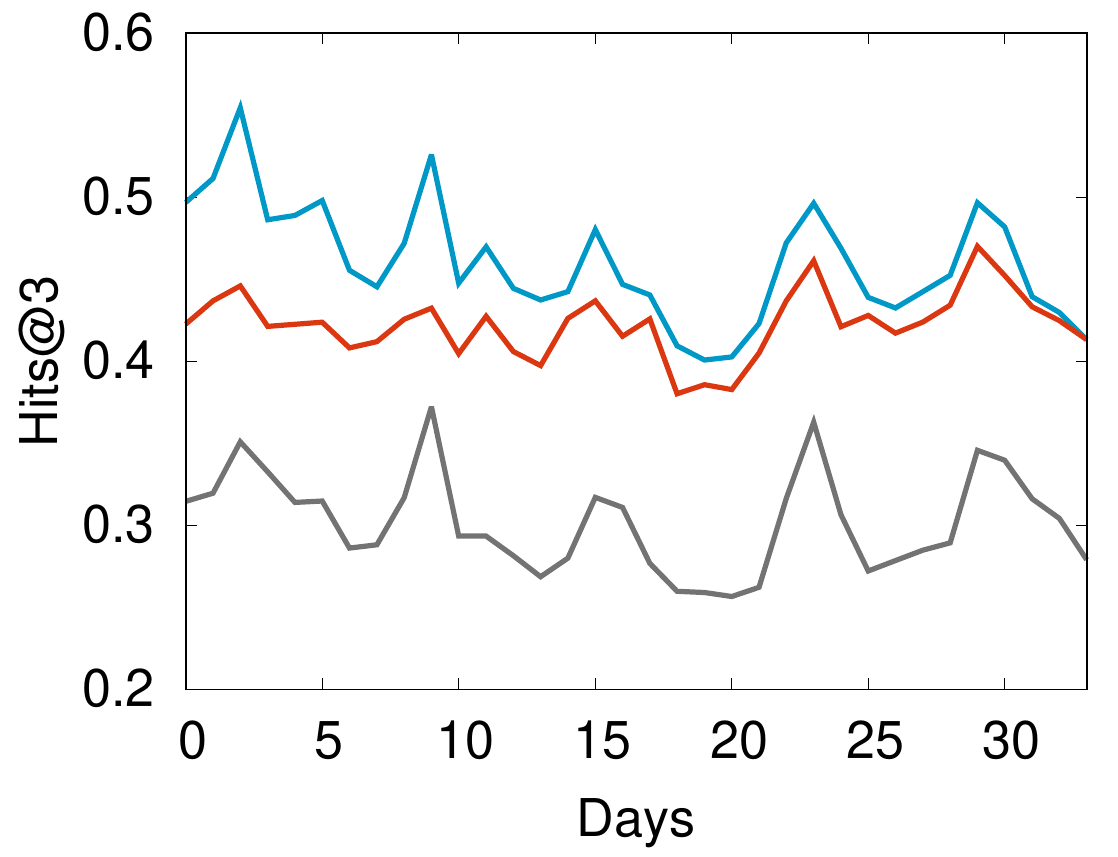}}
    \quad
    \subfloat[GDELT]{\includegraphics[width=0.48\columnwidth]{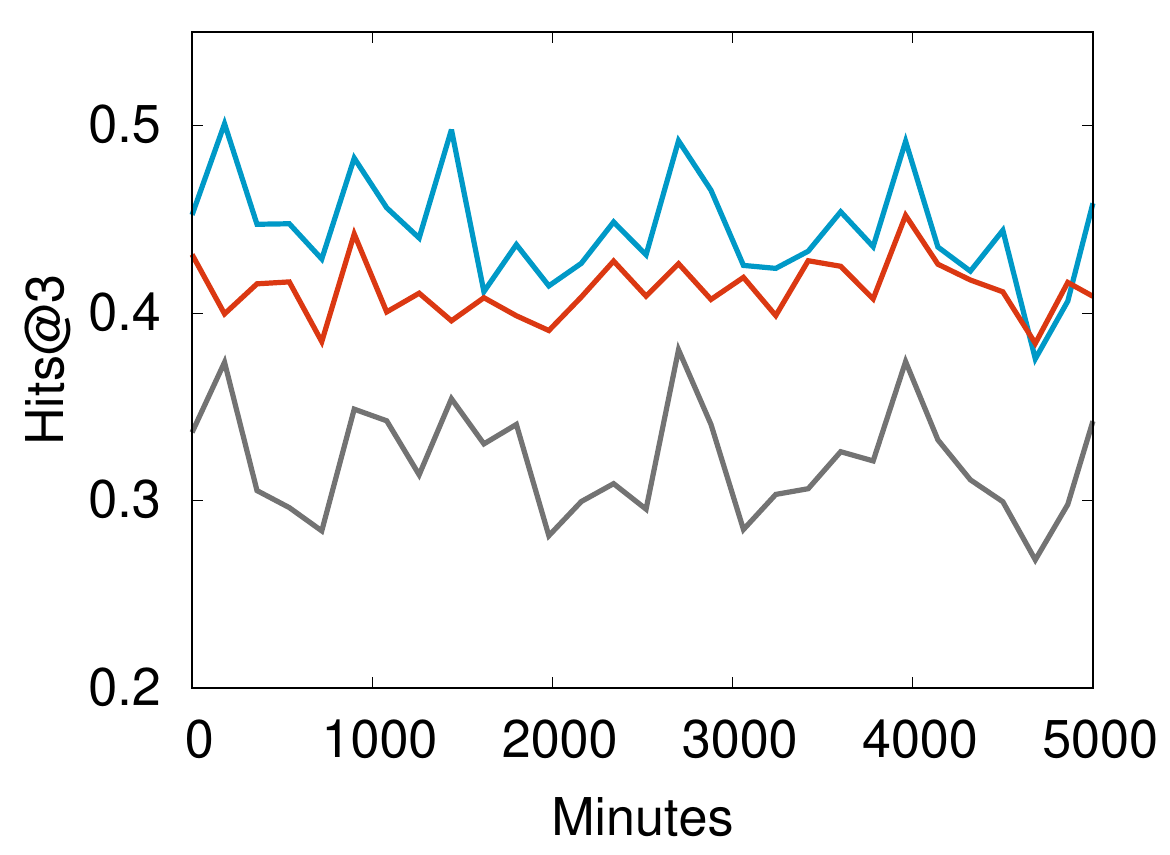}}
    \caption{\textbf{Performance of temporal link prediction over future timestamps with filtered Hits@3.} \method consistently outperforms the baselines. 
    % Results on other datasets are in the appendix.
    }
    \label{fig:timeh3}
\end{figure}

\para{(2) \textit{Temporal Reasoning Methods.}}
We also compare state-of-the-art temporal reasoning methods for knowledge graphs, including Know-Evolve\footnote{*: We found a problematic formulation in Know-Evolve. Details of this issues are discussed in Section~\ref{supp:know} of appendix.}~\citep{Trivedi2017KnowEvolveDT}, TA-DistMult~\citep{GarcaDurn2018LearningSE}, and HyTE~\citep{Dasgupta2018HyTEHT}. 
TA-DistMult and HyTE are for an interpolation task whereas we focus on an extrapolation task. To do this, we assign random values to temporal embeddings that are not observed during training.
To see the effectiveness of our recurrent event encoder, we use encoders of previous work and our MLP decoder as baselines; we compare Know-Evolve, Dyrep~\citep{Trivedi2019DyRepLR}, and GCRN~\citep{Seo2017StructuredSM} combined with our MLP decoder, called Know-Evolve+MLP, DyRep+MLP, and R-GCRN+MLP. The GCRN utilizes Graph Gonvolutional Network~\citep{Kipf2016SemiSupervisedCW}. Instead, we use RGCN~\citep{Schlichtkrull2018ModelingRD} to deal with multi-relational graphs. 

We also compare our method with dynamic methods on homogeneous graphs: dyngraph2vecAE~\citep{goyal2019dyngraph2vec}, tNodeEmbed~\citep{singer2019node}, and EvolveRGCN~\citep{Pareja2019EvolveGCNEG}. 
These methods were proposed to predict interactions at future time on homogeneous graphs. 
Thus, we modified the methods to apply them on multi-relational graph.
% \footnote{Details are provided in Section~\ref{append:base} of appendix.}

\para{(3) \textit{Variants of \method.}} 
To evaluate the importance of different components of \method, we varied our model in different ways:
\method w/o multi-step which does not update history during inference, \method without the aggregator (\method w/o agg.), \method with a mean aggregator (\method w. mean agg.), and \method with an attentive aggregator (\method w. attn agg.). \method w/o agg. takes a zero vector instead of an aggregator.
\method w. GT denotes \method with ground truth history.

Please refer to Section~\ref{append:base} of appendix for detailed experimental settings.
% \footnote{Experimental details of \method and baselines are described in Section~\ref{append:base} of appendix.}
% or interactions during multi-step inference, and thus the model knows all the interactions before the time for testing. 
% It does not update history (or generate a graph) since it already has ground truth history.

\begin{figure}[!tb]
    \centering
        \subfloat[\method with different aggregators]{\includegraphics[width=0.35\columnwidth]{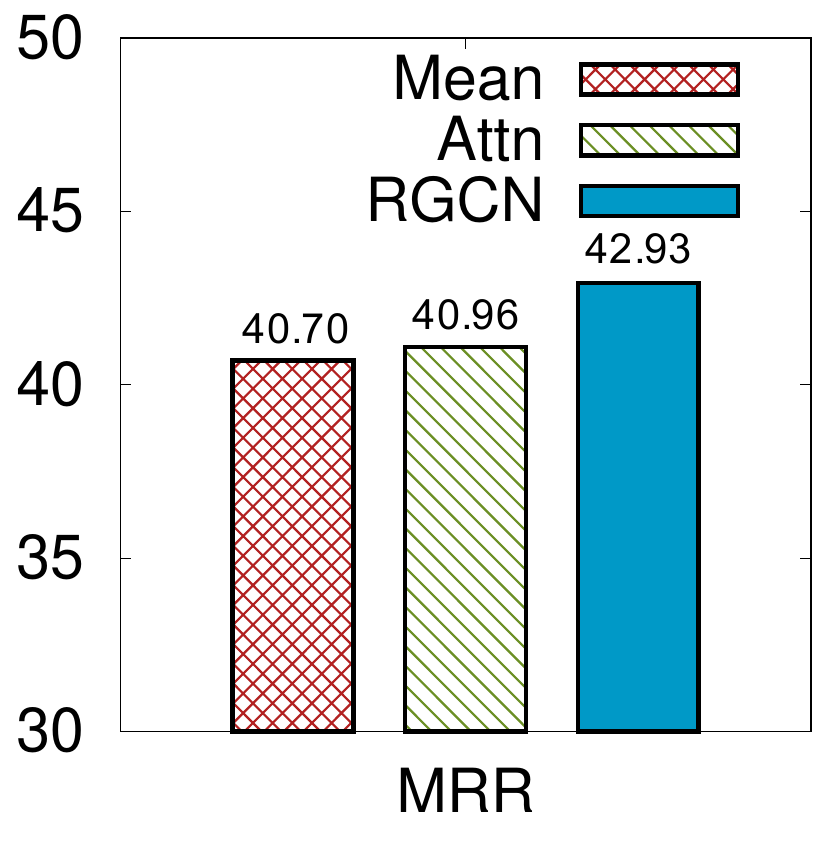} \label{fig:diffaggre}}
        \qquad
        %\hspace{-0.5cm}
        % \subfloat[Effect of global representations]{\includegraphics[width=0.35\columnwidth]{figures/global.pdf} \label{fig:global}}
        % \qquad
        %\hspace{-0.5cm}
        \subfloat[Study of empirical $\prob(s)$ and $\prob(s,r)$]{\includegraphics[width=0.37\columnwidth]{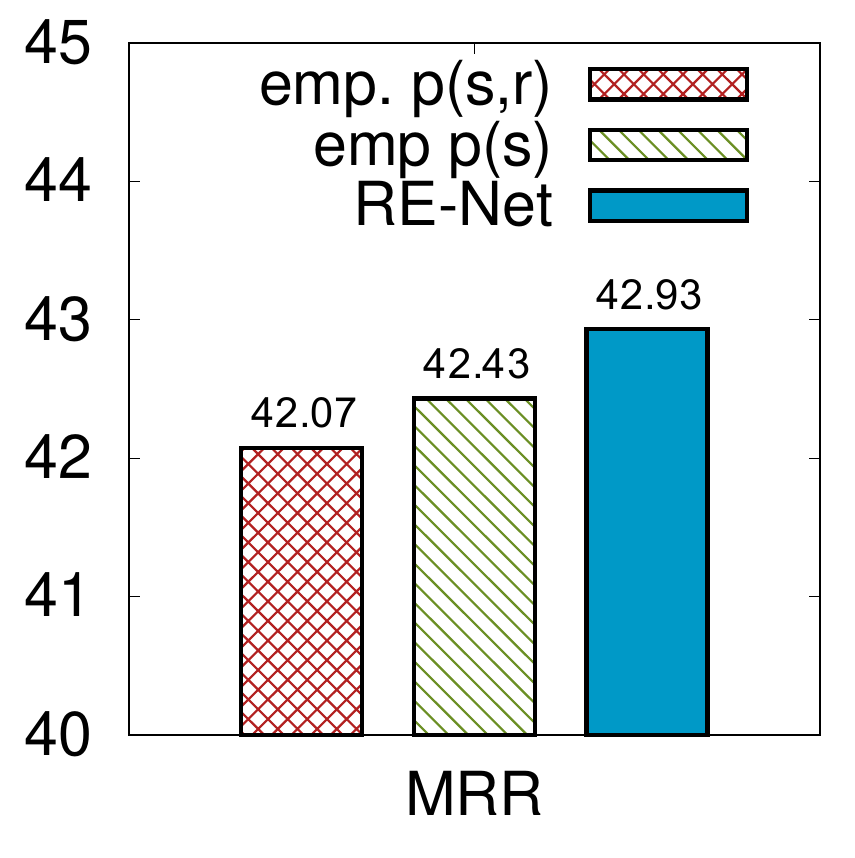}
        \label{fig:empps}}
        % \qquad
        % \subfloat[\# layers of RGCN]{\includegraphics[width=0.36\columnwidth]{figures/layers.pdf} \label{fig:layers}}
        \caption{\textbf{Performance study on model variations.} We study the effects of (a) \method with different aggregators, and (b) empirical $\prob(\bs)$ and $\prob(\bs,\br)$.
        % , and (d) number of RGCN layers in neighborhood aggregation.
        }
        % length of RNN history in recurrent event encoder, and  (d) cutoff position used in entity prediction at inference time.}
        \label{fig:variation}
        % , and (c) number of RGCN layers in neigborhood aggregation.}
\end{figure}

\subsection{Performance Comparison on TKGs.}
We compare our proposed method  with other baselines.
The test results are obtained by averaged metrics (5 runs) over the entire test sets on datasets.

\para{Results on Event-based TKGs.}
Table~\ref{result:filtered} summarizes results on all datasets. 
% Among them, ICEWS18 and GDELT are event-based datasets.
Our proposed \method outperforms all other baselines on ICEWS18 and GDELT.
Static methods underperform compared to our method since they do not consider temporal factors.
Also, \method outperforms all other temporal methods including TA-DistMult, HyTE, and dynamic methods on homogeneous graphs.
Know-Evovle+MLP significantly improves Know-Evolve, which shows effectiveness of our MLP decoder.
However, there is still a large gap from our model, which also indicates effectiveness of our recurrent event encoder. 
% We notice that Know-Evolve and DyRep has a gradient exploding issue on their encoder since their RNN-like structures keep accumulating embedding over time. This issue degrades their performance.
% Graph Convolutional Recurrent Network (GCRN) is not for dynamic and multi-relational graphs, and is not capable of link prediction.
% We modified GCRN to work on dynamic graphs and our problem setting by using RGCN instead of GCN, and our MLP decoder.
% R-GCRN+MLP shows good performance but it does not outperform our method.
R-GCRN+MLP has a similar structure to ours in that it has a recurrent encoder and an RGCN aggregator but it lacks multi-step inference, global information, and the sophisticated modeling for the recurrent encoder. Thus, it underperforms compared to our method.
% These results of the combined models suggest the our recurrent event encoder shows better performance in link prediction.
More importantly, none of the prior temporal methods are capable of multi-step inference, while \method can sequentially infer multi-step events (Details in Section~\ref{exp:ablation}).

\para{Results on Public KGs.}
Previous results have demonstrated the effectiveness of \method on event-based KGs. In Table~\ref{result:filtered} we compare \method with other baselines on the Public KGs  WIKI and YAGO.
Our proposed \method outperforms all other baselines on these datasets. 
In these datasets, baselines show better results than in the event-based TKGs.
This is due to the characteristics of the datasets; they have facts that are valid within a time span. 
However, our proposed method consistently outperforms the static and temporal methods, which implies that \method effectively infers new events using a powerful event encoder and an aggregator, and provides accurate prediction results.

% \subsection{Performance of Prediction over Time.}
\para{Performance of Prediction over Time.}
% Previous results have proved the effectiveness of \method.
Next, we further study performance of \method over time. 
% \meng{Give a background about this section, i.e., why we analyze the performance of prediction over time. For example, previous results have proved the effectiveness of \method in... Next, we further study......}
Figs.~\ref{fig:timeh3} shows the performance comparisons over different time stamps on the ICEWS18, GDELT, WIKI, and YAGO datasets with filtered Hits@3 metrics.
% \method outperforms other baselines with the MRR metric on both datasets. \meng{Where can we see the MRR results?}
% However, \method and ConvE are comparable on the Hits$@3$ metric.
\method consistently outperforms baseline methods for all different time stamps.
Performance of each method fluctuate since testing entities are different at each time step.
We notice that with increasing time steps, the difference between \method and ConvE gets smaller as shown in Fig.~\ref{fig:timeh3}.
This is expected since further future events are harder to predict.
% Thus, we conjecture that the decline of the performance is due to the generation of a long graph sequence.
To estimate the joint probability distribution of events in a distant future, \method needs to generate a long graph sequence.
The quality of generated graphs deteriorates when \method generates a long graph sequence.

\begin{table}[tb!]
\centering
\caption{\textbf{Ablation study on the ICEWS18 and GDELT datasets.}
}
\label{result:aggre}
\resizebox{1\columnwidth}{!}{
\begin{tabular}{l|ccc|ccc}
\cmidrule[1pt](){1-7}
\multicolumn{1}{c}{\multirow{1}{*}{\hspace{-15mm}\vspace{-4.8mm}\textbf{Method}}} & \multicolumn{3}{c}{\textbf{ICEWS18}} & \multicolumn{3}{c}{\textbf{GDELT}} \\ 
\cmidrule(lr){2-4} \cmidrule(lr){5-7}
                        &MRR    &H@3    &H@10   &MRR    &H@3    &H@10     \\

\cmidrule{1-7}
\method w/o agg.        &33.46  &35.98  &46.62  &38.10  &41.26  &51.61 \\
\method w/o multi-step  &40.05  &42.60  &52.92  &38.72  &42.52  &52.78 \\
\method                 &42.93  &45.47  &55.80  &40.42  &43.40  &53.70 \\
\cmidrule{1-7}
\method w. GT           &44.33  &46.83  &57.27  &41.80  &45.71  &56.03\\
\cmidrule[1pt](){1-7}
\end{tabular}}
\end{table}

\subsection{Ablation Study}
\label{exp:ablation}
In this section, we study the effect of variations in \method on the ICEWS18 dataset.
% To evaluate the importance of different components of \method, we varied our model in different ways, measuring the change in performance on the link prediction task on the ICEWS18 dataset. 
We present the results in Tables~\ref{result:filtered}, \ref{result:aggre}, and Fig.~\ref{fig:variation}.

\para{Different Aggregators.}
% We first analyze the effect of different aggregators.
In Table~\ref{result:aggre}, we observe that \method w/o agg. hurts model quality, suggesting that introducing aggregators makes the model capable of dealing with concurrent events and improves performance.
% This suggests that introducing aggregators makes the model capable of dealing with concurrent events and the proposed aggregators improve the prediction performance.
Table~\ref{result:filtered} and Fig.~\ref{fig:diffaggre} show the performance of \method with different aggregators. 
Among them, RGCN aggregator outperforms other aggregators.
This aggregator has the advantage of exploring multi-relational neighbors. 
% not limited to neighbors under the same relation.
Also, \method with an attentive aggregator shows better performance than \method with a mean aggregator, which implies that giving different attention weights to each neighbor helps predictions.

\para{Multi-step Inference.}
In Table~\ref{result:aggre}, we observe that \method outperforms \method w/o multi-step. The latter one does not update history during inference; keeps its last history in the training set. So it is not affected by time stamps. 
Without the multi-step inference, the performance of \method is decreased as is shown.
Also we expect that \method w. GT shows significant improvement when \method uses ground truth of triples at the previous time step which are not allowed in our setup.

% \para{Global Information.}
% We further observe that representations from global graph structures help the predictions. Fig.~\ref{fig:global} shows effectiveness of a representation of global graph structures. 
% The improvement is marginal, but we consider that global representations at different time steps give distinct information beyond local graph structures.

\para{Empirical Probabilities.}
Here, we study the role of $\prob(\bs_{t}|\calE_{t-m:t-1})$ and $\prob(\br_{t}|\bs,\calE_{t-m:t-1})$. 
We denote them as $\prob(\bs)$ and $\prob(\br)$ for brevity. 
$\prob(\bs_t, \br_t| \calE_{t-m:t-1})$ (or simply $\prob(\bs, \br)$) is equivalent to $\prob(\bs) \prob(\br)$.
In Fig~\ref{fig:empps}, emp. $\prob(\bs)$ (or $\prob_e(\bs)$) denotes a model with empirical $\prob(\bs)$, defined as $\prob_e(\bs)=$ (\# of $\bs$-related triples) / (total \# of triples). 
emp. $\prob(\bs,\br)$ (or $\prob_e(\bs,\br)$) denotes a model with $\prob_e(\bs)$ and $\prob_e(\br)$,defined as $\prob_e(\br)=$ (\# of $\br$-related triples) / (total \# of triples). Thus, $\prob_e(\bs,\br) = \prob_e(\bs) \prob_e(\br)$. 
Note that \method use a trained $\prob(\bs)$ and $\prob(\br)$.
The results show that the trained $\prob(\bs)$ and $\prob(\br)$ help \method for multi-step predictions. 
$\prob_e(\bs)$ underperforms \method, and $\prob_e(\bs,\br)=\prob_e(\bs) \prob_e(\br)$ shows the worst performance, suggesting that training each part of the probability in equation~\eqref{eq:prob_event_t} improves performance.

% \para{Additional Analysis.}
% We refer readers to Sections~\ref{append:additional} and \ref{supp:case} of appendix for sensitivity analysis and case study.

%% file: 040related.tex
%Our work is related to previous studies on temporal knowledge graph reasoning, temporal modeling on homogeneous graphs, recurrent graph neural networks, and deep autoregressive models.

\para{Temporal KG Reasoning.}
% Reasoning on temporal knowledge graphs can be deployed under two settings: \textit{interpolation}, which is to make predictions at some time $t$ such that $t_0 \leq t \leq t_T$, and \textit{extrapolation}, which is to predict at future time $t$ such that $t > t_T$. 
% Note that our work focuses on \textit{extrapolation}.
There have been some recent attempts on incorporating temporal information in modeling dynamic knowledge graphs, broadly categorized into two settings - \textit{extrapolation}~\citep{Trivedi2017KnowEvolveDT} and \textit{interpolation}~\citep{GarcaDurn2018LearningSE,leblay2018deriving,Dasgupta2018HyTEHT,goel2020diachronic,Lacroix2020Tensor}.
For the former setting, Know-Evolve~\citep{Trivedi2017KnowEvolveDT} models the occurrence of a fact as a temporal point process. 
% However, this method is built on a problematic formulation when dealing with concurrent events, as shown in Section~\ref{supp:know}.
For the latter setting, several embedding-based methods have been proposed~\citep{GarcaDurn2018LearningSE,leblay2018deriving,Dasgupta2018HyTEHT,goel2020diachronic,Lacroix2020Tensor} to model time information.
They embed the associate into a low dimensional space such as relation embeddings with RNN on the text of time~\citep{GarcaDurn2018LearningSE}, time embeddings~\citep{leblay2018deriving}, temporal hyperplanes~\citep{leblay2018deriving}, diachronic entity embedding~\citep{goel2020diachronic}, and tensor decomposition~\citep{Lacroix2020Tensor}.
However, these models cannot predict future events, as representations of unseen time stamps are unavailable.

% these models do not capture temporal dependency and cannot generalize to unobserved time stamps.
% Our proposed method introduces a recurrent event encoder to capture temporal dependency and is able to predict interactions at unseen time tamps.

\para{Temporal Modeling on Homogeneous Graphs.}
There are attempts on predicting future links on homogeneous graphs~\citep{Pareja2019EvolveGCNEG,goyal2018dyngem,goyal2019dyngraph2vec,zhou2018dynamic,singer2019node}.
Some of the methods try to incorporate and learn graphical structures to predict future links \citep{Pareja2019EvolveGCNEG,zhou2018dynamic,singer2019node}, while other methods predict by reconstructing an adjacency matrix by using an autoencoder \citep{goyal2018dyngem,goyal2019dyngraph2vec}.
These methods seek to predict on single-relational graphs, and are designed to predict future edges in one future step (i.e., for $t+1$).
However, our work focuses on \textit{multi-relational} knowledge graphs and aims for multi-step prediction.

\para{Deep Autoregressive Models.} 
Deep autoregressive models define joint probability distributions as a product of conditionals. DeepGMG~\citep{li2018learning} and GraphRNN~\citep{you2018graphrnn} are deep generative models of graphs and focus on generating static homogeneous graphs where there is only a single type of edge. In contrast to these studies, our work focuses on generating heterogeneous graphs, in which multiple types of edges exist, and thus our problem is more challenging. 
% Our proposed work is to design a tractable autorgressive model to generate heterogeneous graphs. 
To the best of our knowledge, this is the first paper to formulate the structure inference (prediction) problem for temporal, multi-relational (knowledge) graphs in an autoregressive fashion.

%% file: 060conclusion.tex
% In this work, we studied the sequential graph generation on temporal knowledge graphs.
% In this work, we frame link prediction in TKG as a sequence prediction problem. 
To tackle the extrapolation problem, we proposed Recurrent Event Network (\method) to model temporal, multi-relational, and concurrent interactions between entities. 
% A recurrent event encoder in \method summarizes information of the past event sequences, and a neighborhood aggregator collects the information of concurrent.
\method defines the joint probability of all events, and thus is capable of inferring graphs in a sequential manner.
% We tested the proposed model on a link prediction task on temporal knowledge graphs. 
The experiment revealed that \method outperforms all the static and temporal methods and our extensive analysis shows its strength.
% We study various kinds of aggregators, and evaluate our proposed work with different aggregators. Our extensive experiments show the effectiveness of \method on predicting unseen relationships over time on two TKG datasets. 
Interesting future work includes developing a fast and efficient version of \method, and modeling lasting events and performing inference on the long-lasting graph structures.

%% file: appendix.tex
% \begin{table*}[h]
%     \centering
%     % \small
%     \caption{Dataset Statistics. }
%     \scalebox{1}{
%     \begin{tabular}{ccccccc}
%          \hline
%          Data & $N_{train}$ &$N_{valid}$ & $N_{test}$ & $N_{ent}$ & $N_{rel}$ & Time granularity\\
%          \hline
%          GDELT    & 1,734,399 &238,765 &305,241 &7,691   &240& 15 mins\\
%          ICEWS18  & 373,018 &45,995 &49,545 & 23,033 & 256 & 24 hours\\
%          ICEWS14  & 323,895 & - & 341,409 & 12,498 & 260 & 24 hours\\
%          \hline
%          WIKI     & 539,286 & 67,538& 63,110& 12,554&   24& 1 year\\
%          YAGO     & 161,540 & 19,523& 20,026 &   10,623	&10 & 1 year\\
%          \hline
%     \end{tabular}}
%     \label{tab:dataset}
% \end{table*}

\section{Recurrent Neural Network}

\label{append:eventseq}
We define a recurrent event encoder based on RNN as follows:
\begin{equation*}
	\small \bh_{t}(\bs,\br) = \textrm{RNN}(g(\teN_{t}(\bs)), \bH_t, \bh_{t-1}(\bs,\br)).
\end{equation*}

We use Gated Recurrent Units~\citep{Cho2014LearningPR} as RNN:

\small
\begin{align*}
    \boldsymbol{a}_{t} &= [\es: \er: g(\teN_{t}(\bs)): \bH_t]\\
    \boldsymbol{z}_t &= \sigma(\bW_z \boldsymbol{a}_{t} + \boldsymbol{U}_z \bh_{t-1})\\
    \boldsymbol{r}_t &= \sigma(\bW_r \boldsymbol{a}_{t} + \boldsymbol{U}_r \bh_{t-1})\\
    \bh_t &= (1-\boldsymbol{z}_t)\circ \bh_{t-1} + \boldsymbol{z}_t \circ \tanh(\bW_h \boldsymbol{a}_{t} + \boldsymbol{U}_h (\boldsymbol{r}_t \circ \bh_{t-1})),
\end{align*}
\normalsize
where $:$ is concatenation, $\sigma(\cdot)$ is an activation function, and $\circ$ is a Hadamard operator.
The input is a concatenation of four vectors: subject embedding, object embedding, aggregation of neighborhood representations, and global information vector ($\es, \er, g(\teN_{t}(\bs)), \bH_t$).
$\bh_t(s)$ and $\bH_t$ are similarly defined. For $\bh_t(s)$, a concatenation of subject embedding, aggregation of neighborhood representations, and global information vector ($\es, g(\teN_t(s)), \bH_t$) is input. For $\bH_t$, aggregation of the whole graph representations $g(G_t)$ is input.

\section{Details of RGCN Aggregator}
\label{sec:rgcn}
The RGCN aggregator is defined as follows:
\begin{equation}
\label{eq:rgcn}
    \small g(\teN^{(\bs)}_t)= \bbh_s^{(l+1)} = \sigma\Big(\sum_{\br\in R}\sum_{\bo \in \teN_t^{(\bs,\br)}} \frac{1}{c_s} \bW_r^{(l)} \bbh_\bo^{(l)} + \bW_0^{(l)} \bbh_s^{(l)}\Big), 
\end{equation}
where initial hidden representations for each node ($\bbh_o^{(0)}$) are set to trainable embedding vectors ($\eo$) for each node and $c_s$ is a normalizing factor. Detailed

Basically, each relation can derive a local graph structure between entities, which further yield a message on each entity by aggregating information from neighbors, i.e., {\small  $\sum_{\bo \in \teN_t^{(\bs,\br)}} \allowbreak \frac{1}{c_s} \bW_r^{(l)} \bbh_\bo^{(l)}$}. The overall message on each entity is further computed by aggregating all the relation-specific messages, i.e., {\small $\sum_{\br\in R}\sum_{\bo \in \teN_t^{(\bs,\br)}}\allowbreak \frac{1}{c_s} \bW_r^{(l)} \bbh_\bo^{(l)}$}. Finally, the aggregator {\small  $g(\teN^{(\bs)}_t)$} is defined by combining both the overall message and information from past steps, i.e., {\small $\bW_0^{(l)} \bbh_s^{(l)}$}.

To distinguish weights between different relations, we adopt independent weight matrices $\{ \bW_\br^{(l)} \}$ for each relation $\br$. Furthermore, the aggregator collects representations of multi-hop neighbors by introducing multiple layers of the neural network with each layer indexed by $l$.
The number of layers determines the depth to which the node reaches to aggregate information from its local neighborhood. 

The major issue of this aggregator is that the number of parameters grows rapidly with the number of relations.
In practice, this can easily lead to overfitting on rare relations and models of very large size.
Thus, we adopt the block-diagonal decomposition~\citep{Schlichtkrull2018ModelingRD}, where each relation-specific weight matrix is decomposed into a block-diagonal by decomposing into low-dimensional matrices.
$\bW_r^{(l)}$ in equation \eqref{eq:rgcn} is defined as a block diagonal matrix, $\text{diag}(\mathbf{A}_{1r}^{(l)},..., \mathbf{A}_{Br}^{(l)})$ where {\small $\mathbf{A}_{kr}^{(l)} \in \mathbb{R}^{(d^{(l+1)}/B) \times (d^{(l)}/B)}$} and $B$ is the number of basis matrices.
The block decomposition reduces the number of parameters and helps prevent overfitting.

\section{Computational Complexity Analysis.}
\label{sec:complexity}
We analyze the time complexity of the graph generation process in Algorithm~\ref{alg:renet}. Computing $\prob(\bs_{t}|\allowbreak\calE_{t-m:t-1})$ (equation \eqref{eq:prob3}) takes $O(|E|Lm)$, where $|E|$ is the maximum number of triples among $\{G_{t-m},...,G_{t-1}\}$, $L$ is the number of layers of aggregation, and $m$ is the number of the past time steps since we unroll $m$ time steps in RNN. From this probability, we sample $M$ number of subjects $\bs$. Computing $\prob(\bs_t,\br_t,\bo_t|\calE_{t-m:t-1})$ in equation \eqref{eq:prob_event_t} takes $O(D^L m)$, where $D$ is the maximum degree of entities. To get probabilities of all possible triples given sampled subjects, it requires $O(M|R||O|D^Lm)$ where $|R|$ is the total number of relations and $|O|$ is the total number of entities. Thus, the time complexity for generating one graph is $O(|E|Lm+M|R||O|(D^L m + \log k))$ where $k$ is the cutoff number for picking top-$k$ triples. The time complexity is linear to the number of entities and relations, and the number of sampled $\bs$.

\begin{table*}[t]
    \centering
    % \small
    \caption{\textbf{Dataset Statistics.}}
    \scalebox{0.89}{
    \begin{tabular}{ccccccc}
         \hline
         Data & $N_{train}$ &$N_{valid}$ & $N_{test}$ & $N_{ent}$ & $N_{rel}$ & Time gap\\
         \hline
         GDELT    & 1,734,399 &238,765 &305,241 &7,691   &240& 15 mins\\
         ICEWS18  & 373,018 &45,995 &49,545 & 23,033 & 256 & 24 hours\\
         ICEWS14  & 323,895 & - & 341,409 & 12,498 & 260 & 24 hours\\
         \hline
         WIKI     & 539,286 & 67,538& 63,110& 12,554&   24& 1 year\\
         YAGO     & 161,540 & 19,523& 20,026 &   10,623	&10 & 1 year\\
         \hline
    \end{tabular}}
    \label{tab:dataset}
\end{table*}

\section{Detailed Experimental Settings}
\label{append:base}

\para{Datasets.}
\label{append:dataset}
We use five datasets: 1) three event-based temporal knowledge graphs and 2) two knowledge graphs where temporally associated facts have meta-facts as $(\bs,\br,\bo, [t_s, t_e])$ where $t_s$ is the starting time point and $t_e$ is the ending time point. 
The first group of graphs includes Integrated Crisis Early Warning System (ICEWS18 ~\citep{boschee2015icews} and ICEWS14~\citep{Trivedi2017KnowEvolveDT}), and Global Database of Events, Language, and Tone (GDELT)~\citep{leetaru2013gdelt}.
% ICEWS is collected from 1/1/2018 to 10/31/2018, and GDELT is from 1/1/2018 to 1/31/2018.
% Both datasets include records of events that include two actors, action type and timestamp of the event.
The second group of graphs includes WIKI~\citep{leblay2018deriving} and YAGO~\citep{Mahdisoltani2014YAGO3AK}.
% WIKI and YAGO datasets have temporally associated facts $(\bs,\br,\bo, [t_s , t_e ])$. 
Dataset statistics are described on Table~\ref{tab:dataset}, where $N_{train}$, $N_{valid}$, and $N_{test}$ are the numbers of train set, valid set, and test set, respectively. $N_{ent}$ and $N_{rel}$ are the numbers of entities and relations. The time gap represents time granularity between adjacent events.

% We use five datasets: 1) three event-based temporal knowledge graphs (ICEWS18, ICEWS14, and GDELT), and 2) two knowledge graphs (WIKI and YAGO). 
% The first group of graphs includes Integrated Crisis Early Warning System (ICEWS18)~\citep{boschee2015icews} and Global Database of Events, Language, and Tone (GDELT)~\citep{leetaru2013gdelt}.
ICEWS18 is collected from 1/1/2018 to 10/31/2018, ICEWS14 is from 1/1/2014 to 12/31/2014, and GDELT is from 1/1/2018 to 1/31/2018.
The ICEWS14 is from \citep{Trivedi2017KnowEvolveDT}. We didn't use their version of the GDELT dataset since they didn't release the dataset.
% Both datasets include records of events that include two actors, action type and timestamp of the event.
% The second group of graphs includes WIKI~\cite{leblay2018deriving} and YAGO~\cite{Mahdisoltani2014YAGO3AK}.

WIKI and YAGO datasets have temporally associated facts $(\bs,\br,\bo, \allowbreak [t_s, t_e])$.
We preprocess the datasets such that each fact is converted to $\{(\bs,\br,\bo, t_s), (\bs,\br,\bo,t_\bs+1_t),...,(\bs,\br,\bo,t_e)\}$ where $1_t$ is a unit time to ensure each fact has a sequence of events. Noisy events of early years are removed (before 1786 for WIKI and 1830 for YAGO). 
% We also noticed there are missing begin dates and end dates in YAGO. Some relationships lasting for a period are preprocessed as lasting for the rest of missing time. Other relationships which happen only once are preprocessed as one time events.

The difference between the first group and the second group is that facts happen multiple times (even periodically) on the first group (event-based knowledge graphs) while facts last long time but are not likely to occur multiple times in the second group.

\para{Model details of \method.}
We use Gated Recurrent Units \citep{Cho2014LearningPR} as our recurrent event encoder, where the length of history is set as $m = 10$ which means saving past 10 event sequences.
If the events related to $\bs$ are sparse, we check the previous time steps until we get $m$ previous time steps related to the entity $\bs$.
We pretrain the parameters related to equations \ref{eq:prob3} and \ref{eq:globalh} due to the large size of training graphs. 
% We freeze the parameters during learning parameters for equations \ref{eq:prob} and \ref{eq:prob2}. 
We use a multi-relational aggregator to compute $\bH_t$. The aggregator provides hidden representations for each node and we max-pool over all hidden representations to get $\bH_t$.
% We use a multi-layer perceptron with one hidden layer followed by a softmax layer for $f$ in \eqref{eq:prob}.
% Thus, 10-steps past observations (or predictions) are fed into them to predict next entities.
% We preprocessed the dataset such that each subject has an 10-lengtth object history and an each object has a 10-length subject history.
% During inference, we do not use ground truth history, but predicted history for a fair comparison.
We apply teacher forcing for model training over historical data, i.e., we use the ground truth rather than the model's own prediction as the input of the next time step during training.
At inference time, \method performs multi-step prediction across the time stamps in dev and test sets.
In each time step, we sample 1000 $(=M)$ number of subjects and save top-1000 $(=k)$ triples to use them as a generated graph .
% In other words, we predict future entities based on the previous predictions.
We set the size of entity/relation embeddings to be 200 and embedding of unobserved embeddings are randomly initialized. 
We use two-layer RGCN in the RGCN aggregator with block dimension $2 \times 2$.
% We choose top-10 entities to save as history at each inference.
The model is trained by the Adam optimizer~\citep{kingma2014adam}.
We set $\lambda_1$ to 0.1, the learning rate to $0.001$ and the weight decay rate to 0.00001.
All experiments were done on GeForce GTX 1080 Ti.

% \begin{table}[tb!]
% \caption{Performance comparison on the ICEWS14 dataset with the filtered setting. 
% }
% \label{result:icews14}
% \centering
% \resizebox{0.8\columnwidth}{!}{
% \begin{tabular}{ll|ccc}
% \cmidrule[1pt](){2-5}
%     \multicolumn{2}{c}{\multirow{2}{*}{\vspace{-2.2mm}\hspace{-11mm}\textbf{Method}}} &\multicolumn{3}{c}{\textbf{ICEWS14}}\\ \cmidrule(lr){3-5} 
%     &                   &MRR    &H@3    &H@10  \\ 
% \cmidrule{2-5}
% \multirow{4}{*}{\rotatebox{90}{\hspace*{-6pt}Static}} 
% &DistMult&19.06  &22.00  &36.41  \\
% &R-GCN&26.31  &30.43  &45.34 \\
% &ConvE&40.73  &43.92  &54.35 \\
% &RotatE&29.56  &32.92  &42.68 \\
% \cmidrule[0.6pt](){2-5}
% \multirow{12}{*}{\rotatebox{90}{\hspace*{-6pt}Temporal}}
% &TA-DistMult&20.78  &22.80  &35.26 \\ 
% &HyTE&11.48  &13.04  &22.51   \\ 
% \cmidrule{2-5}
% &dyngraph2vecAE &10.83  &12.70  &15.02    \\
% &tNodeEmbed&17.84  &20.16  &32.88    \\
% &EvolveRGCN&17.01  &18.97  &32.58   \\
% &Know-Evolve*&1.42   &1.37   &1.43   \\ 
% &Know-Evolve+MLP &22.89  &26.68  &38.57 \\ 
% &DyRep+MLP &24.61  &28.87  &39.34  \\ 
% &R-GCRN+MLP&36.77  &40.15  &52.33   \\ 
% \cmidrule{2-5}
% &\method w. mean agg.&43.79  &47.34  &57.47\\
% &\method w. attn agg.&43.94  &47.85  &57.91 \\
% &\method  &\textbf{45.71} &\textbf{49.06} &	\textbf{59.12}  \\
% \cmidrule[1pt](){2-5}
% \end{tabular}}
% \end{table}

\begin{table*}[tb!]
\caption{\textbf{Performance comparisons with raw metrics.} We observe our method outperforms all other methods.
}
\label{result:raw}
\resizebox{1\textwidth}{!}{
\begin{tabular}{ll|ccc|ccc|ccc|ccc|ccc}
\cmidrule[1pt](){2-17}
    \multicolumn{2}{c}{\multirow{2}{*}{\vspace{-2.2mm}\hspace{-10mm}\textbf{Method}}} & \multicolumn{3}{c}{\textbf{ICEWS18}} & \multicolumn{3}{c}{\textbf{GDELT}} & \multicolumn{3}{c}{\textbf{ICEWS14}} &\multicolumn{3}{c}{\textbf{WIKI}}& \multicolumn{3}{c}{\textbf{YAGO}} \\ \cmidrule(lr){3-5} \cmidrule(lr){6-8} \cmidrule(lr){9-11} \cmidrule(lr){12-14} \cmidrule(lr){15-17}
    &                   &MRR    &H@3    &H@10   &MRR    &H@3    &H@10   &MRR    &H@3    &H@10   &MRR    &H@3    &H@10   &MRR    &H@3    &H@10  \\ 
\cmidrule{2-17}
\multirow{4}{*}{\rotatebox{90}{\hspace*{-6pt}Static}} 
&DistMult               &13.86  &15.22  &31.26  &8.61	&8.27   &17.04  &9.72   &10.09  &22.53  &27.96  &32.45  &39.51  &44.05  &49.70  &59.94   \\
&R-GCN                  &15.05  &16.49  &29.00  &12.17  &12.37  &20.63  &15.03  &16.12  &31.47  &13.96  &15.75  &22.05  &27.43  &31.24  &44.75   \\
&ConvE                  &22.56  &25.41  &41.67  &18.43  &19.57  &32.25  &21.64  &23.16  &38.37  &26.41  &30.36	&39.41  &41.31  &47.10	&59.67 \\
&RotatE                 &11.63  &12.31  &28.03  &3.62   &2.26   &8.37   &9.79   &9.37   &22.24  &26.08  &31.63  &38.51  &42.08  &46.77  &59.39   \\
\cmidrule[0.6pt](){2-17}
\multirow{12}{*}{\rotatebox{90}{\hspace*{-6pt}Temporal}}
&TA-DistMult            &15.62  &17.09  &32.21  &10.34  &10.44  &21.63  &11.29  &11.60  &23.71  &26.44  &31.36  &38.97  &44.98  &50.64  &61.11   \\
&HyTE                   &7.41   &7.33   &16.01  &6.69   &7.57   &19.06  &7.72   &7.94   &20.16  &25.40  &29.16  &37.54  &14.42  &39.73  &46.98   \\ 
\cmidrule{2-17}
&dyngraph2vecAE         &1.36   &1.54   &1.61   &4.53   &1.87   &1.87   &6.95   &8.17   &12.18  &2.67   &2.75   &3.00   &0.81   &0.74   &0.76    \\
% &DynTriad               &xxx    &xxx    &xxx    &xxx    &xxx    &xxx    &xxx    &xxx    &xxx    &xxx    &xxx    &xxx    &xxx    &xxx    &xxx    \\
&tNodeEmbed             &7.21   &7.64   &15.75  &12.97  &12.61  &21.22  &13.36  &13.13  &24.31  &8.86   &10.11  &16.36  &3.82   &3.88   &8.07    \\
&EvolveRGCN             &10.31  &10.52  &23.65  &6.54   &5.64   &15.22  &8.32   &7.64   &18.81  &27.19  &31.35  &38.13  &40.50  &45.78  &55.29    \\
&Know-Evolve*           &0.11   &0.00   &0.47   &0.11   &0.02   &0.10   &0.05   &0.00   &0.10   &0.03   &0      &0.04   &0.02   &0	    &0.01    \\ 
&Know-Evolve+MLP        &7.41   &7.87   &14.76  &15.88  &15.69  &22.28  &16.81  &18.63  &29.20  &10.54  &13.08	&20.21  &5.23   &5.63   &10.23   \\ 
&DyRep+MLP              &7.82   &7.73   &16.33  &16.25  &16.45  &23.86  &17.54  &19.87  &30.34  &10.41  &12.06  &20.93  &4.98   &5.54   &10.19   \\ 
&R-GCRN+MLP             &23.46  &26.62  &41.96  &18.63  &19.80  &32.42  &21.39  &23.60  &38.96  &28.68  &31.44  &38.58  &43.71  &48.53  &56.98   \\ 
\cmidrule{2-17}
&\method w. mean agg.   &25.45  &29.27  &44.31  &19.03  &20.20  &33.32  &22.73  &25.47  &41.48  &30.19  &32.94  &40.57  &46.33  &52.49  &61.21   \\
&\method w. attn agg.   &25.76  &29.56  &44.86  &19.35  &20.42  &33.55  &23.18  &25.98  &41.95  &30.25  &30.12  &40.86  &46.56  &52.56  &61.35   \\
&\method                &26.62  &30.27  &45.57  &19.60  &20.56  &33.89  &23.85  &14.63  &42.58  &30.87  &33.55  &41.27  &46.81  &52.71  &61.93 \\
\cmidrule[1pt](){2-17}
\end{tabular}}
\end{table*}

\para{Experimental Settings for Baseline Methods.}
In this section, we provide detailed settings for baselines.
We use implementations of DistMult\footnote{\scriptsize https://github.com/jimmywangheng/knowledge\_representation\_pytorch}.
We implemented TA-DistMult based on the implementation of Distmult.
% Results of TTransE, TA-TransE, DistMult and TA-DistMult were produced by adding model implementations using the same framework. 
For TA-DistMult, We use temporal tokens with the vocabulary of year, month and day on the ICEWS dataset and the vocabulary of year, month, day, hour and minute on the GDELT dataset. 
% The activation function of hidden states and cells in LSTM for time aware models were linear activation functions. 
We use use a binary cross-entropy loss for DistMult and TA-DistMult.
We validate the embedding size among 100 and 200. 
We set the batch size to 1024, margin to 1.0, negative sampling ratio to 1, and use the Adam optimizer. 

% We use the implementation of ComplEx\footnote{https://github.com/thunlp/OpenKE}~\cite{Han2018OpenKEAO}.
% % Results of ComplEx and RESCAL were produced by Han's implementations\footnote{https://github.com/thunlp/OpenKE} (\cite{Han2018OpenKEAO}).
% We validate the embedding size among 50, 100 and 200. 
% The batch size is 100, the margin is 1.0, and the negative sampling ratio is 1.
% % Other settings were default settings in the implementations. 
% % The batch number was set to 100, the ratio of negative samples to 1, the margin to 1.0, the learning rate to 0.1.
% We use the Adagrad optimizer.

We use the implementation of HyTE\footnote{\scriptsize https://github.com/malllabiisc/HyTE}.
% Results of HyTE were produced by the author's implementation\footnote{https://github.com/malllabiisc/HyTE}.
% We added the raw setting when loading data. 
% We adopted early stopping method which stops learning when the MRR values didn't improve.
% We added validation part and stopped learning when the MRR values didn't improve. 
% Since most timestamps of both ICEWS18 and GDELT have more than 300 events, 
We use every timestamp as a hyperplane. 
% We used the default setting -- 
The embedding size is set to 128, the negative sampling ratio to 5, and margin to 1.0. 
We use time agnostic negative sampling (TANS) for entity prediction, and the Adam optimizer.

We use the codes for ConvE\footnote{\scriptsize https://github.com/TimDettmers/ConvE} and use implementation by Deep Graph Library\footnote{\scriptsize https://github.com/dmlc/dgl/tree/master/examples/pytorch/rgcn}. 
Embedding sizes are 200 for both methods.
We use 1 to all negative sampling for ConvE and use 10 negative sampling ratio for RGCN, and use the Adam optimizer for both methods.
We use the codes for Know-Evolve\footnote{\scriptsize https://github.com/rstriv/Know-Evolve}.
For Know-Evolve, we fix the issue in their codes. Issues are described in Section~\ref{supp:know}.
We follow their default settings.

We use the code for RotatE\footnote{\scriptsize https://github.com/DeepGraphLearning/KnowledgeGraphEmbedding}. The hidden layer/embedding size is set to 100, and batch size 256; other values follow the best values for the larger FB15K dataset configurations supplied by the author. The author reports filtered metrics only, so we added the implementation of the raw setting.

% \subsection{Comparisons with Dynamic Methods.}

\para{Experimental Settings for Dynamic Methods.}
We compare our method with dynamic methods on homogeneous graphs: dyngraph2vecAE~\citep{goyal2019dyngraph2vec}, tNodeEmbed~\citep{singer2019node}, and EvolveGCN-O~\citep{Pareja2019EvolveGCNEG}. 
These methods were proposed to predict interactions at a future time on homogeneous graphs, while our proposed method is for predicting interactions on multi-relational graphs (or knowledge graphs). Furthermore, those methods predict links at one future time stamp, whereas our method seeks to predict interactions at multiple future time stamps.
We modified some methods to apply them on multi-relational graphs as follows.
We adopt R-GCN~\citep{Schlichtkrull2018ModelingRD} for EvolveGCN-O and call it EvolveRGCN. 
We convert knowledge graphs into homogeneous graphs for dyngraph2vecAE. 
% We remove relations and build an adjacency matrix for dyngraph2vecAE. 
The idea of this method is to reconstruct an adjacency matrix using an auto-encoder and regard it as a future adjacency matrix.
If we keep relations, relation-specific adjacency matrices will be extremely sparse; the method learns to reconstruct near-zero adjacency matrices.
% We modified dyngraph2vecAE to use it on multi-relational graphs. We build relation-specific adjacency matrices $\boldsymbol{A}_\br$ so the method can predict $\boldsymbol{A}_\br^t$ from $\{\boldsymbol{A}_\br^1,...,\boldsymbol{A}_\br^{t-1}\}$. We share parameters of an auto-encoder across relations. 
tNodeEmbed is a temporal method on homogeneous graphs. To use this on multi-relational graphs, we first train entity embeddings with DistMult and set these as initial embeddings for entities in tNodeEmbed. Also we give entity embeddings as input to LSTM of tNodeEmbed. We concatenate output of LSTM and relation embeddings to predict objects. 
We did not modified other methods since it is not trivial to extend the methods.

\begin{figure}[!tb]
    \centering
    {\includegraphics[width=0.8\columnwidth]{figures/legend.pdf}
    % \vspace{-3mm}
    }\\
    \subfloat[WIKI]{\includegraphics[width=0.46\columnwidth]{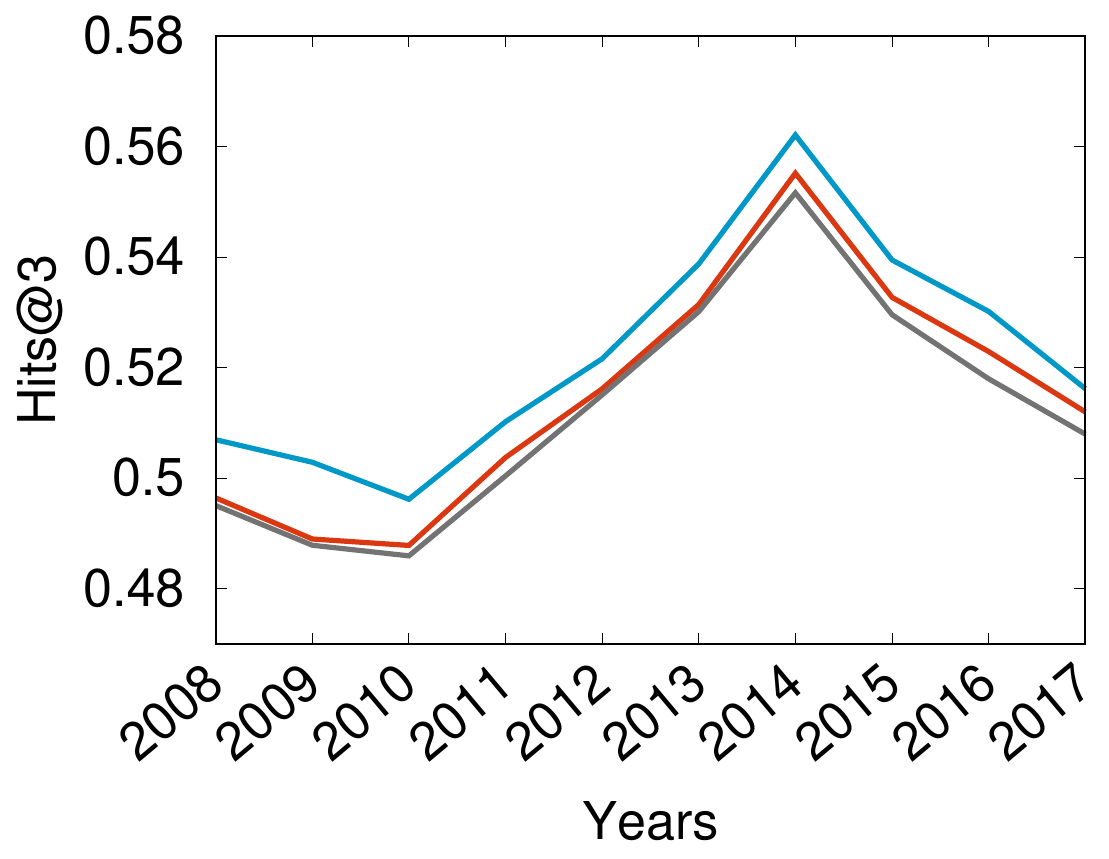}}
    \quad
    \subfloat[YAGO]{\includegraphics[width=0.46\columnwidth]{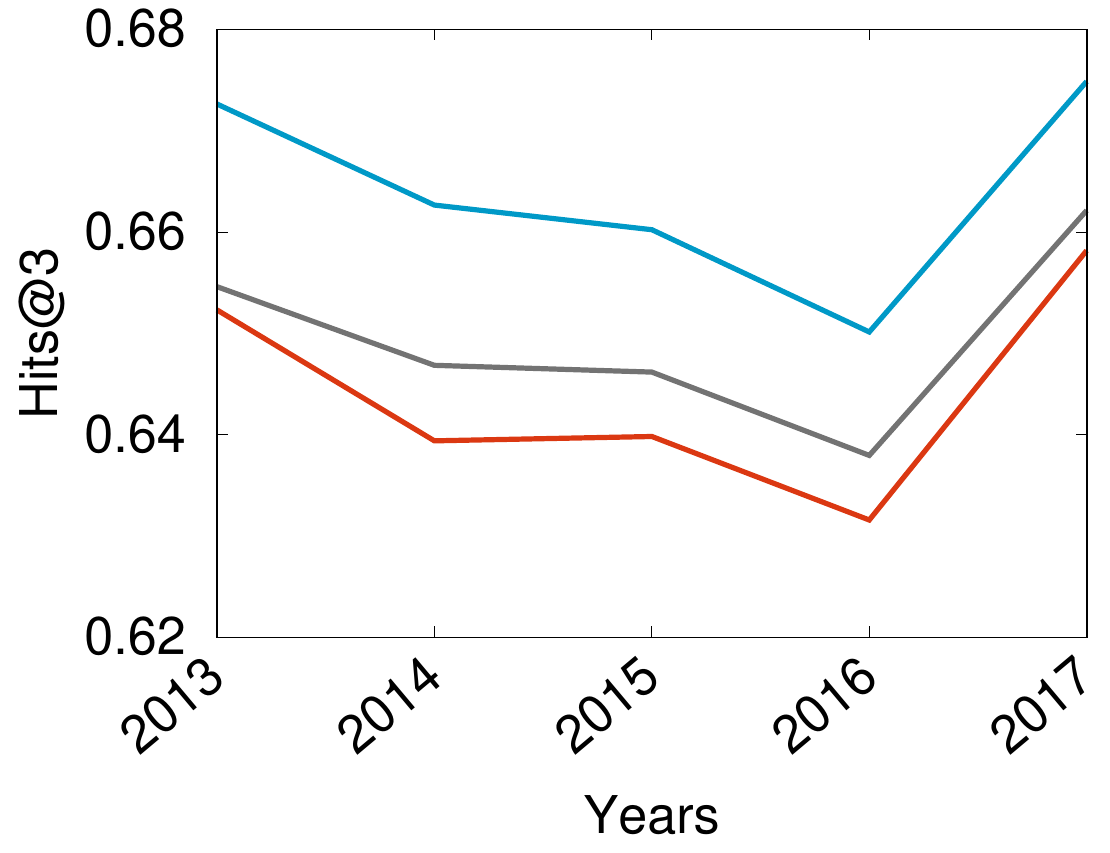}}
    \caption{\textbf{Performance of temporal link prediction over future timestamps with filtered Hits@3.} \method consistently outperforms the baselines.
    }
    \label{fig:timeh3_2}
\end{figure}

\section{Additional Experiments}
\label{append:additional}
% \subsection{Results on ICEWS14}
% Table~\ref{result:icews14} shows the results on the ICEWS14 dataset with a filtered metric.

\subsection{Results with Raw Metrics}
Table~\ref{result:raw} shows the performance comparison on ICEWS18, GDELT, ICEWS14 with raw settings.
Our proposed \method outperforms all other baselines. 
% \textbf{Performance of Prediction over Time.}
% Figs.~\ref{fig:timemrr} shows the performance comparisons over different time stamps on the ICEWS18, GDELT, WIKI and YAGO datasets with filtered MRR.
% Our proposed \method consistently outperform baselines over time.

\begin{figure}[!t]
% \vspace{-0.4cm}
    \centering
        \subfloat[Length of past history]{\includegraphics[width=0.43\columnwidth]{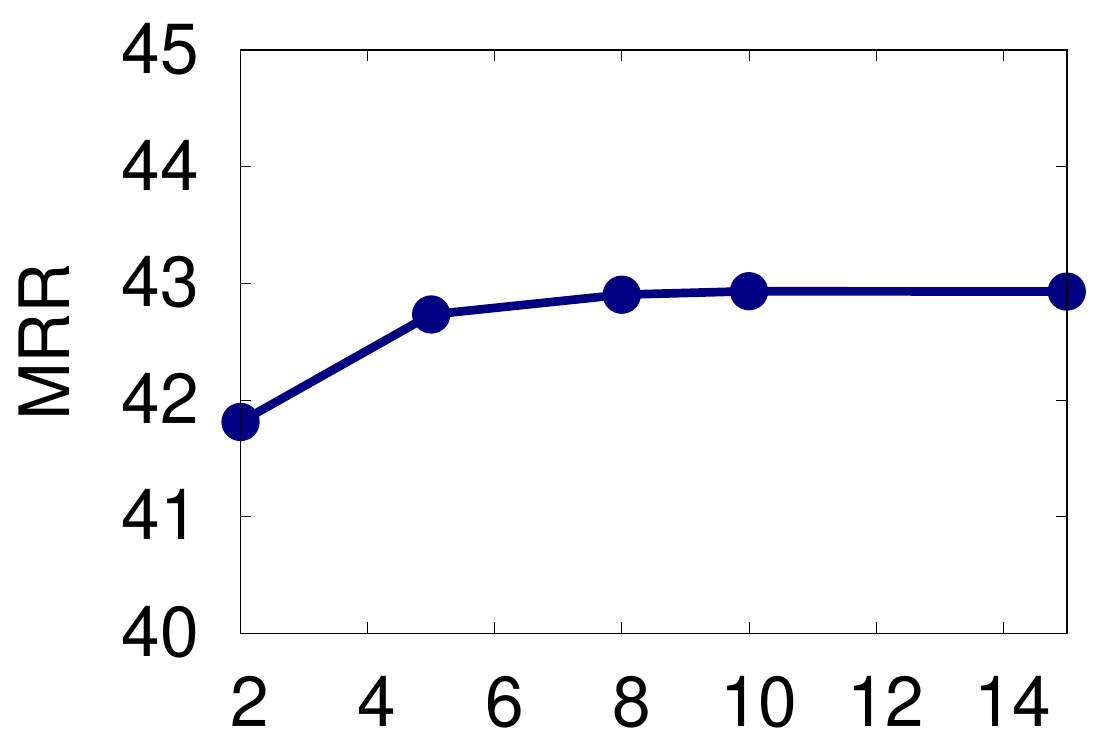} \label{fig:history}}
        \quad
        \subfloat[Cutoff position $k$]{\includegraphics[width=0.45\columnwidth]{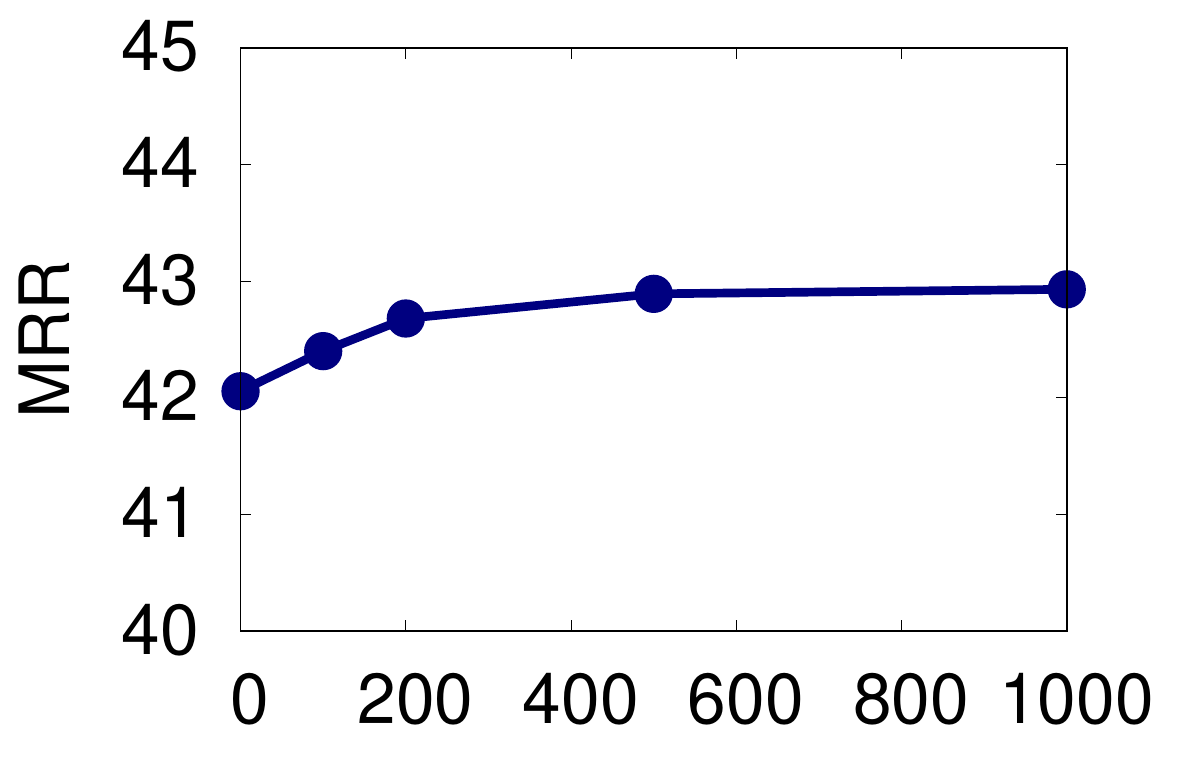} 
        \label{fig:topk}}
        \quad
        \subfloat[\# layers of RGCN]{\includegraphics[width=0.36\columnwidth]{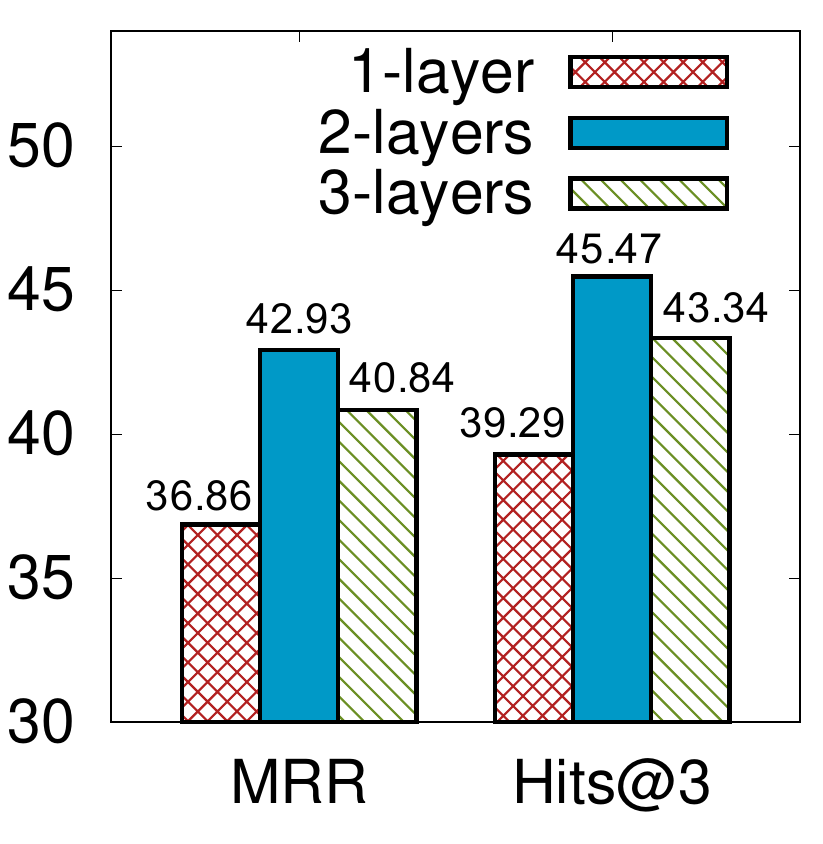} \label{fig:layers}}
        \qquad
        \subfloat[Effect of global representations]{\includegraphics[width=0.35\columnwidth]{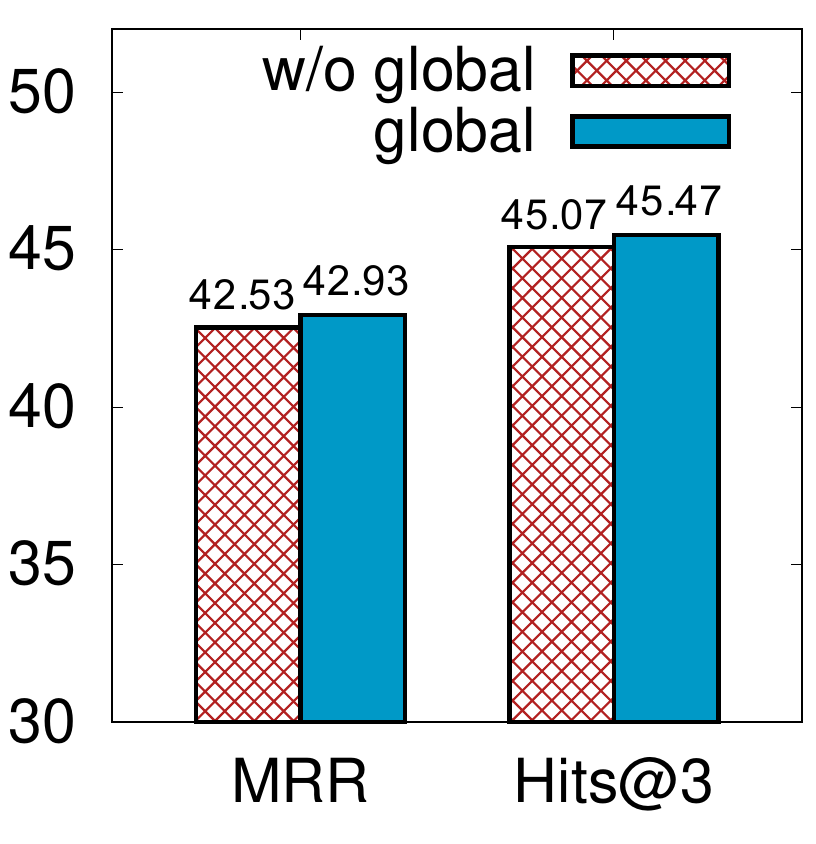} \label{fig:global}}
        % \\
        % % \qquad
        % \subfloat[\# layers of RGCN]{\includegraphics[width=0.4\columnwidth]{figures/layers.pdf} \label{fig:layers}}
        % \vspace{-0.3cm}
        \caption{\textbf{Parameter sensitivity on \method.}
        We study the effects of (a) length of RNN history in event sequence encoder, and (b) cutoff position at inference time, (c) number of RGCN layers in neighborhood aggregation, and (d) effect of the global representation from a global graph structure.
        }
        \label{fig:variation2}
\end{figure}

\subsection{Sensitivity Analysis}
\label{sec:sens}
In this section, we study the parameter sensitivity of \method including the length of history for the event encoder, cutoff position k for events to generate a graph, the number of layers of the RGCN aggregator, and effect of the global representation from a global graph structure.
We report the performance change of \method on the ICEWS18 dataset by varying the hyper-parameters (Figs.~\ref{fig:variation2} and \ref{fig:layers}).

\para{Length of Past History in Recurrent Event Encoder.}
The recurrent event encoder takes the sequence of past interactions up to $m$ graph sequences or previous histories. 
Fig.~\ref{fig:history} shows the performance with various lengths of past histories.
When \method uses longer histories, MRR is getting higher.
However, the MRR is not likely to go higher when the length of history is 5 and over.
% This implies that long history does not make big differences.

\para{Cut-off Position $k$ at Inference.} 
To generate a graph at each time, we cut off top-$k$ triples on ranking results.
In Fig.~\ref{fig:topk},
% shows the performance by choosing different cutoff position $k$. 
when $k$ is 0, \method does not generate graphs for estimating $\prob(G_{t+\Delta t}|G_{:t})$, i.e., \method performs single-step predictions, and it shows the lowest result.
When $k$ is larger, the performance is getting higher and it is saturated after 500.
We notice that the conditional distribution $\prob(G_{t+\Delta t}|G_{:t})$ can be approximated by $\prob(G_{t+\Delta t}|\allowbreak \hat{G}_{t+1:t+\Delta t -1},G_{:t})$ by using a larger cutoff position.

\para{Layers of RGCN Aggregator.}
% We examine the number of layers in the RGCN aggregator. 
The number of layers in the aggregator means the depth to which the node reaches.
Fig.~\ref{fig:layers} shows the performance according to different numbers of layers of RGCN.
2-layered RGCN improves the performance considerably compared to 1-layered RGCN since 2-layered RGCN aggregates more information.
However, \method with 3-layered RGCN underperforms \method with 2-layered RGCN. 
We conjecture that the bigger parameter space leads to overfitting.

\para{Global Information.}
We further observe that representations from global graph structures help the predictions. Fig.~\ref{fig:global} shows effectiveness of a representation of global graph structures. 
The improvement is marginal, but we consider that global representations at different time steps give distinct information beyond local graph structures.

\begin{figure*}[tb!]
    \centering
    \includegraphics[width=0.85\linewidth]{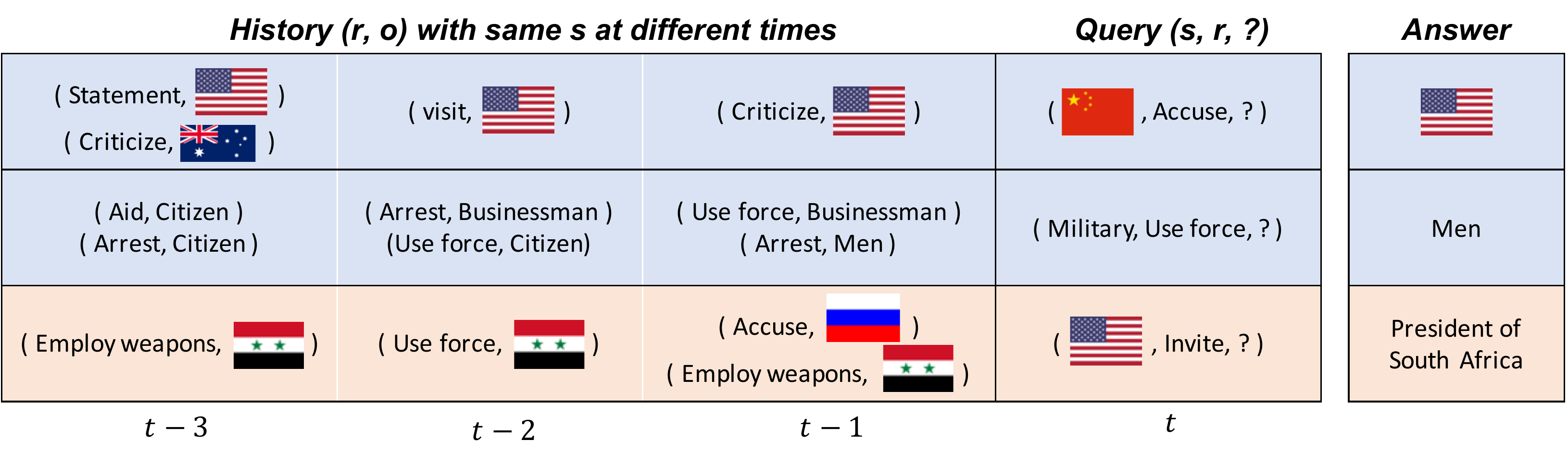}
    % \vspace{-0.3cm}
    \caption{\textbf{Case study of \method's predictions.} \method's predictions depend on interaction histories. Interaction histories are categorized into three cases: (1) consistent interactions with an object, (2) a specific temporal pattern, and (3) irrelevant history. \method achieves good performances on the first two cases, and poor performances on the third case.
    % \wj{add percentage.}
    % \xiang{with some space left, can you expand and enrich the case study figure?}
    }
    \label{fig:case}
\end{figure*}

\section{Case Study}
\label{supp:case}
In this section, we study \method's predictions. Its predictions depend on interaction histories. We categorize histories into three cases: (1) consistent interactions with an object, (2) a specific temporal pattern, and (3) irrelevant history (Fig.~\ref{fig:case}).
\method can learn (1) and (2) cases, so it achieves high performances.
For the first case, \method can predict the answer because it consistently interacts with an object. However, static methods are prone to predicting different entities which are observed under relation "Accuse" in training set. 
The second case shows specific temporal patterns on relations: ( Arrest, $o$ ) $\rightarrow$ ( Use force, $o$ ). Without knowing this pattern, one method might predict ``Businessman" instead of ``Men". \method is able to learn these temporal patterns so it can predict the second case. 
Lastly, the third case shows irrelevant history to the answer and the history is not helpful to predictions. \method fails to predict the third case.

\section{Implementation Issues of Know-Evolve}
\label{supp:know}

We found a problematic formulation in the Know-Evolve model and codes.
The intensity function (equation 3 in \citep{Trivedi2017KnowEvolveDT}) is defined as 
$\lambda_r^{s,r}(t|\bar{t})=f(g_r^{s,r}(\bar{t}))(t-\bar{t})$, where $g(\cdot)$ is a score function, $t$ is current time, and $\bar{t}$ is the most recent time point when either subject or object entity was involved in an event.
This intensity function is used in inference to rank entity candidates.
However, they don't consider concurrent event at the same time stamps, and thus $\bar{t}$ will become $t$ after one event.
For example, we have events $e_1=(s,r,o_1, t_1), e_2=(s, r, o_2, t_1)$.
After $e_1$, $\bar{t}$ will become $t$ (subject $s$'s most recent time point), and thus the value of intensity function for $e_2$ will be 0.
This is problematic in inference since if $t=\bar{t}$, then the intensity function will always be 0 regardless of entity candidates. 
In inference, all object candidates are ranked by the intensity function.
% So we are given $(s,r)$ and $t=\bar{t}$.
But all intensity scores for all candidates will be 0 since $t=\bar{t}$, which means all candidates have the same 0 score.
In their code, they give the highest ranks (first rank) for all entities including the ground truth object in this case.
Thus, we fixed their code for a fair comparison; we give an average rank to entities who have the same scores.

%% file: 0-main.bbl
\begin{thebibliography}{33}
\expandafter\ifx\csname natexlab\endcsname\relax\def\natexlab#1{#1}\fi

\bibitem[{Bahdanau et~al.(2015)Bahdanau, Cho, and
  Bengio}]{Bahdanau2015NeuralMT}
Dzmitry Bahdanau, Kyunghyun Cho, and Yoshua Bengio. 2015.
\newblock Neural machine translation by jointly learning to align and
  translate.
\newblock \emph{CoRR}, abs/1409.0473.

\bibitem[{Bordes et~al.(2013)Bordes, Usunier, Garc{\'i}a-Dur{\'a}n, Weston, and
  Yakhnenko}]{Bordes2013TranslatingEF}
Antoine Bordes, Nicolas Usunier, Alberto Garc{\'i}a-Dur{\'a}n, Jason Weston,
  and Oksana Yakhnenko. 2013.
\newblock Translating embeddings for modeling multi-relational data.
\newblock In \emph{NIPS}.

\bibitem[{Boschee et~al.(2015)Boschee, Lautenschlager, O'Brien, Shellman,
  Starz, and Ward}]{boschee2015icews}
Elizabeth Boschee, Jennifer Lautenschlager, Sean O'Brien, Steve Shellman, James
  Starz, and Michael Ward. 2015.
\newblock Icews coded event data.
\newblock \emph{Harvard Dataverse}, 12.

\bibitem[{Cho et~al.(2014)Cho, van Merrienboer, Çaglar G{\"u}lçehre,
  Bahdanau, Bougares, Schwenk, and Bengio}]{Cho2014LearningPR}
Kyunghyun Cho, Bart van Merrienboer, Çaglar G{\"u}lçehre, Dzmitry Bahdanau,
  Fethi Bougares, Holger Schwenk, and Yoshua Bengio. 2014.
\newblock Learning phrase representations using rnn encoder-decoder for
  statistical machine translation.
\newblock In \emph{EMNLP}.

\bibitem[{Dasgupta et~al.(2018)Dasgupta, Ray, and
  Talukdar}]{Dasgupta2018HyTEHT}
Shib~Sankar Dasgupta, Swayambhu~Nath Ray, and Partha Talukdar. 2018.
\newblock Hyte: Hyperplane-based temporally aware knowledge graph embedding.
\newblock In \emph{EMNLP}.

\bibitem[{Dettmers et~al.(2018)Dettmers, Minervini, Stenetorp, and
  Riedel}]{Dettmers2018Convolutional2K}
Tim Dettmers, Pasquale Minervini, Pontus Stenetorp, and Sebastian Riedel. 2018.
\newblock Convolutional 2d knowledge graph embeddings.
\newblock In \emph{AAAI}.

\bibitem[{Garc{\'i}a-Dur{\'a}n et~al.(2018)Garc{\'i}a-Dur{\'a}n, Dumancic, and
  Niepert}]{GarcaDurn2018LearningSE}
Alberto Garc{\'i}a-Dur{\'a}n, Sebastijan Dumancic, and Mathias Niepert. 2018.
\newblock Learning sequence encoders for temporal knowledge graph completion.
\newblock In \emph{EMNLP}.

\bibitem[{Goel et~al.(2020)Goel, Kazemi, Brubaker, and
  Poupart}]{goel2020diachronic}
Rishab Goel, Seyed~Mehran Kazemi, Marcus Brubaker, and Pascal Poupart. 2020.
\newblock Diachronic embedding for temporal knowledge graph completion.
\newblock In \emph{Thirty-Fourth AAAI Conference on Artificial Intelligence}.

\bibitem[{Goyal et~al.(2019)Goyal, Chhetri, and Canedo}]{goyal2019dyngraph2vec}
Palash Goyal, Sujit~Rokka Chhetri, and Arquimedes Canedo. 2019.
\newblock dyngraph2vec: Capturing network dynamics using dynamic graph
  representation learning.
\newblock \emph{Knowledge-Based Systems}, page 104816.

\bibitem[{Goyal et~al.(2018)Goyal, Kamra, He, and Liu}]{goyal2018dyngem}
Palash Goyal, Nitin Kamra, Xinran He, and Yan Liu. 2018.
\newblock Dyngem: Deep embedding method for dynamic graphs.
\newblock \emph{arXiv preprint arXiv:1805.11273}.

\bibitem[{Kazemi et~al.(2019)Kazemi, Goel, Jain, Kobyzev, Sethi, Forsyth, and
  Poupart}]{kazemi2019relational}
Seyed~Mehran Kazemi, Rishab Goel, Kshitij Jain, Ivan Kobyzev, Akshay Sethi,
  Peter Forsyth, and Pascal Poupart. 2019.
\newblock Relational representation learning for dynamic (knowledge) graphs: A
  survey.
\newblock \emph{arXiv preprint arXiv:1905.11485}.

\bibitem[{Kingma and Ba(2014)}]{kingma2014adam}
Diederik~P Kingma and Jimmy Ba. 2014.
\newblock Adam: A method for stochastic optimization.
\newblock \emph{arXiv preprint arXiv:1412.6980}.

\bibitem[{Kipf and Welling(2016)}]{Kipf2016SemiSupervisedCW}
Thomas~N. Kipf and Max Welling. 2016.
\newblock Semi-supervised classification with graph convolutional networks.
\newblock \emph{CoRR}, abs/1609.02907.

\bibitem[{Korkmaz et~al.(2015)Korkmaz, Cadena, Kuhlman, Marathe, Vullikanti,
  and Ramakrishnan}]{korkmaz2015combining}
Gizem Korkmaz, Jose Cadena, Chris~J Kuhlman, Achla Marathe, Anil Vullikanti,
  and Naren Ramakrishnan. 2015.
\newblock Combining heterogeneous data sources for civil unrest forecasting.
\newblock In \emph{Proceedings of the 2015 IEEE/ACM International Conference on
  Advances in Social Networks Analysis and Mining 2015}, pages 258--265.

\bibitem[{Lacroix et~al.(2020)Lacroix, Obozinski, and
  Usunier}]{Lacroix2020Tensor}
Timothée Lacroix, Guillaume Obozinski, and Nicolas Usunier. 2020.
\newblock Tensor decompositions for temporal knowledge base completion.
\newblock In \emph{International Conference on Learning Representations}.

\bibitem[{Leblay and Chekol(2018)}]{leblay2018deriving}
Julien Leblay and Melisachew~Wudage Chekol. 2018.
\newblock Deriving validity time in knowledge graph.
\newblock In \emph{Companion of the The Web Conference 2018 on The Web
  Conference 2018}, pages 1771--1776. International World Wide Web Conferences
  Steering Committee.

\bibitem[{Leetaru and Schrodt(2013)}]{leetaru2013gdelt}
Kalev Leetaru and Philip~A Schrodt. 2013.
\newblock Gdelt: Global data on events, location, and tone, 1979--2012.
\newblock In \emph{ISA annual convention}, volume~2, pages 1--49. Citeseer.

\bibitem[{Li et~al.(2018)Li, Vinyals, Dyer, Pascanu, and
  Battaglia}]{li2018learning}
Yujia Li, Oriol Vinyals, Chris Dyer, Razvan Pascanu, and Peter Battaglia. 2018.
\newblock Learning deep generative models of graphs.
\newblock \emph{arXiv preprint arXiv:1803.03324}.

\bibitem[{Mahdisoltani et~al.(2014)Mahdisoltani, Biega, and
  Suchanek}]{Mahdisoltani2014YAGO3AK}
Farzaneh Mahdisoltani, Joanna~Asia Biega, and Fabian~M. Suchanek. 2014.
\newblock Yago3: A knowledge base from multilingual wikipedias.
\newblock In \emph{CIDR}.

\bibitem[{Morstatter et~al.(2019)Morstatter, Galstyan, Satyukov, Benjamin,
  Abeliuk, Mirtaheri, Hossain, Szekely, Ferrara, Matsui, Steyvers, Bennet,
  Budescu, Himmelstein, Ward, Beger, Catasta, Sosic, Leskovec, Atanasov,
  Joseph, Sethi, and Abbas}]{ijcai2019SAGE}
Fred Morstatter, Aram Galstyan, Gleb Satyukov, Daniel Benjamin, Andres Abeliuk,
  Mehrnoosh Mirtaheri, KSM~Tozammel Hossain, Pedro Szekely, Emilio Ferrara,
  Akira Matsui, Mark Steyvers, Stephen Bennet, David Budescu, Mark Himmelstein,
  Michael Ward, Andreas Beger, Michele Catasta, Rok Sosic, Jure Leskovec, Pavel
  Atanasov, Regina Joseph, Rajiv Sethi, and Ali Abbas. 2019.
\newblock Sage: A hybrid geopolitical event forecasting system.
\newblock In \emph{Proceedings of the Twenty-Eighth International Joint
  Conference on Artificial Intelligence, {IJCAI-19}}. International Joint
  Conferences on Artificial Intelligence Organization.

\bibitem[{Muthiah et~al.(2015)Muthiah, Huang, Arredondo, Mares, Getoor, Katz,
  and Ramakrishnan}]{Muthiah2015PlannedPM}
Sathappan Muthiah, Bert Huang, Jaime Arredondo, David Mares, Lise Getoor,
  Graham Katz, and Naren Ramakrishnan. 2015.
\newblock Planned protest modeling in news and social media.
\newblock In \emph{AAAI}.

\bibitem[{Pareja et~al.(2020)Pareja, Domeniconi, Chen, Ma, Suzumura, Kanezashi,
  Kaler, Schardl, and Leiserson}]{Pareja2019EvolveGCNEG}
Aldo Pareja, Giacomo Domeniconi, Jie Chen, Tengfei Ma, Toyotaro Suzumura,
  Hiroki Kanezashi, Tim Kaler, Tao~B. Schardl, and Charles~E. Leiserson. 2020.
\newblock {EvolveGCN}: Evolving graph convolutional networks for dynamic
  graphs.
\newblock In \emph{Proceedings of the Thirty-Fourth AAAI Conference on
  Artificial Intelligence}.

\bibitem[{Phillips et~al.(2017)Phillips, Dowling, Shaffer, Hodas, and
  Volkova}]{Phillips2017UsingSM}
Lawrence Phillips, Chase Dowling, Kyle Shaffer, Nathan~Oken Hodas, and Svitlana
  Volkova. 2017.
\newblock Using social media to predict the future: A systematic literature
  review.
\newblock \emph{ArXiv}, abs/1706.06134.

\bibitem[{Schlichtkrull et~al.(2018)Schlichtkrull, Kipf, Bloem, van~den Berg,
  Titov, and Welling}]{Schlichtkrull2018ModelingRD}
Michael~Sejr Schlichtkrull, Thomas~N. Kipf, Peter Bloem, Rianne van~den Berg,
  Ivan Titov, and Max Welling. 2018.
\newblock Modeling relational data with graph convolutional networks.
\newblock In \emph{ESWC}.

\bibitem[{Seo et~al.(2017)Seo, Defferrard, Vandergheynst, and
  Bresson}]{Seo2017StructuredSM}
Youngjoo Seo, Micha{\"e}l Defferrard, Pierre Vandergheynst, and Xavier Bresson.
  2017.
\newblock Structured sequence modeling with graph convolutional recurrent
  networks.
\newblock In \emph{ICONIP}.

\bibitem[{Singer et~al.(2019)Singer, Guy, and Radinsky}]{singer2019node}
Uriel Singer, Ido Guy, and Kira Radinsky. 2019.
\newblock Node embedding over temporal graphs.
\newblock \emph{arXiv preprint arXiv:1903.08889}.

\bibitem[{Sun et~al.(2019)Sun, Deng, Nie, and Tang}]{sun2019rotate}
Zhiqing Sun, Zhi-Hong Deng, Jian-Yun Nie, and Jian Tang. 2019.
\newblock Rotate: Knowledge graph embedding by relational rotation in complex
  space.
\newblock \emph{arXiv preprint arXiv:1902.10197}.

\bibitem[{Theis et~al.(2015)Theis, Oord, and Bethge}]{theis2015note}
Lucas Theis, A{\"a}ron van~den Oord, and Matthias Bethge. 2015.
\newblock A note on the evaluation of generative models.
\newblock \emph{arXiv preprint arXiv:1511.01844}.

\bibitem[{Trivedi et~al.(2017)Trivedi, Dai, Wang, and
  Song}]{Trivedi2017KnowEvolveDT}
Rakshit Trivedi, Hanjun Dai, Yichen Wang, and Le~Song. 2017.
\newblock Know-evolve: Deep temporal reasoning for dynamic knowledge graphs.
\newblock In \emph{ICML}.

\bibitem[{Trivedi et~al.(2019)Trivedi, Farajtabar, Biswal, and
  Zha}]{Trivedi2019DyRepLR}
Rakshit Trivedi, Mehrdad Farajtabar, Prasenjeet Biswal, and Hongyuan Zha. 2019.
\newblock Dyrep: Learning representations over dynamic graphs.
\newblock In \emph{ICLR 2019}.

\bibitem[{Yang et~al.(2015)Yang, tau Yih, He, Gao, and
  Deng}]{Yang2015EmbeddingEA}
Bishan Yang, Wen tau Yih, Xiaodong He, Jianfeng Gao, and Li~Deng. 2015.
\newblock Embedding entities and relations for learning and inference in
  knowledge bases.
\newblock \emph{CoRR}, abs/1412.6575.

\bibitem[{You et~al.(2018)You, Ying, Ren, Hamilton, and
  Leskovec}]{you2018graphrnn}
Jiaxuan You, Rex Ying, Xiang Ren, William Hamilton, and Jure Leskovec. 2018.
\newblock Graphrnn: Generating realistic graphs with deep auto-regressive
  models.
\newblock In \emph{International Conference on Machine Learning}, pages
  5694--5703.

\bibitem[{Zhou et~al.(2018)Zhou, Yang, Ren, Wu, and Zhuang}]{zhou2018dynamic}
Lekui Zhou, Yang Yang, Xiang Ren, Fei Wu, and Yueting Zhuang. 2018.
\newblock Dynamic network embedding by modeling triadic closure process.
\newblock In \emph{Thirty-Second AAAI Conference on Artificial Intelligence}.

\end{thebibliography}
